\documentclass[10pt,twocolumn,letterpaper]{article}
\usepackage{cvpr}
\usepackage{multirow}
\usepackage{booktabs} 
\usepackage{amsmath}  
\usepackage{graphicx}
\usepackage{cuted}
\usepackage{capt-of}
\usepackage{algorithm}
\usepackage{algpseudocode}
\usepackage{etoc}
\usepackage{gensymb}
\usepackage{makecell}
\usepackage[accsupp]{axessibility}
\definecolor{cvprblue}{rgb}{0.21,0.49,0.74}
\usepackage[pagebackref,breaklinks,colorlinks,allcolors=cvprblue]{hyperref}
\definecolor{bgcolor}{RGB}{140, 60, 120}
\newcommand{\icon}{\raisebox{-5pt}{\includegraphics[width=0.040\textwidth]{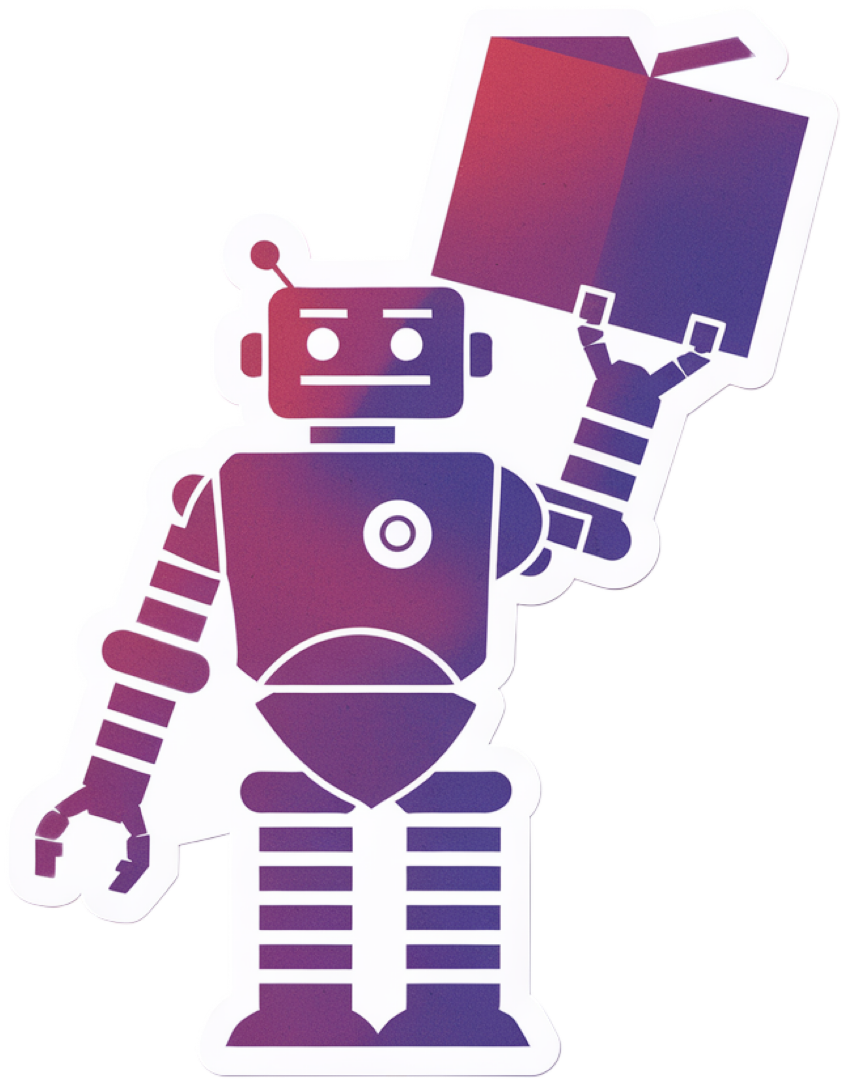}}}
\newcommand{\name}[0]{\textsc{InterMimic}\xspace}

\title{\icon \textcolor{bgcolor}{\name}: Towards Universal Whole-Body Control for Physics-Based Human-Object Interactions}

\author{Sirui Xu$^{1}$ \quad Hung Yu Ling$^{2}$ \quad
Yu-Xiong Wang$^{1\dag}$ \quad
Liang-Yan Gui$^{1\dag}$\\
$^{1}$ University of Illinois Urbana-Champaign \quad $^{2}$ Electronic Arts\\
$^{\dag}$ Equal Advising\\
\small\url{https://sirui-xu.github.io/InterMimic}}
\begin{document}
\maketitle
\begin{strip}\centering
\vspace{-3.2em}
\includegraphics[width=\textwidth]{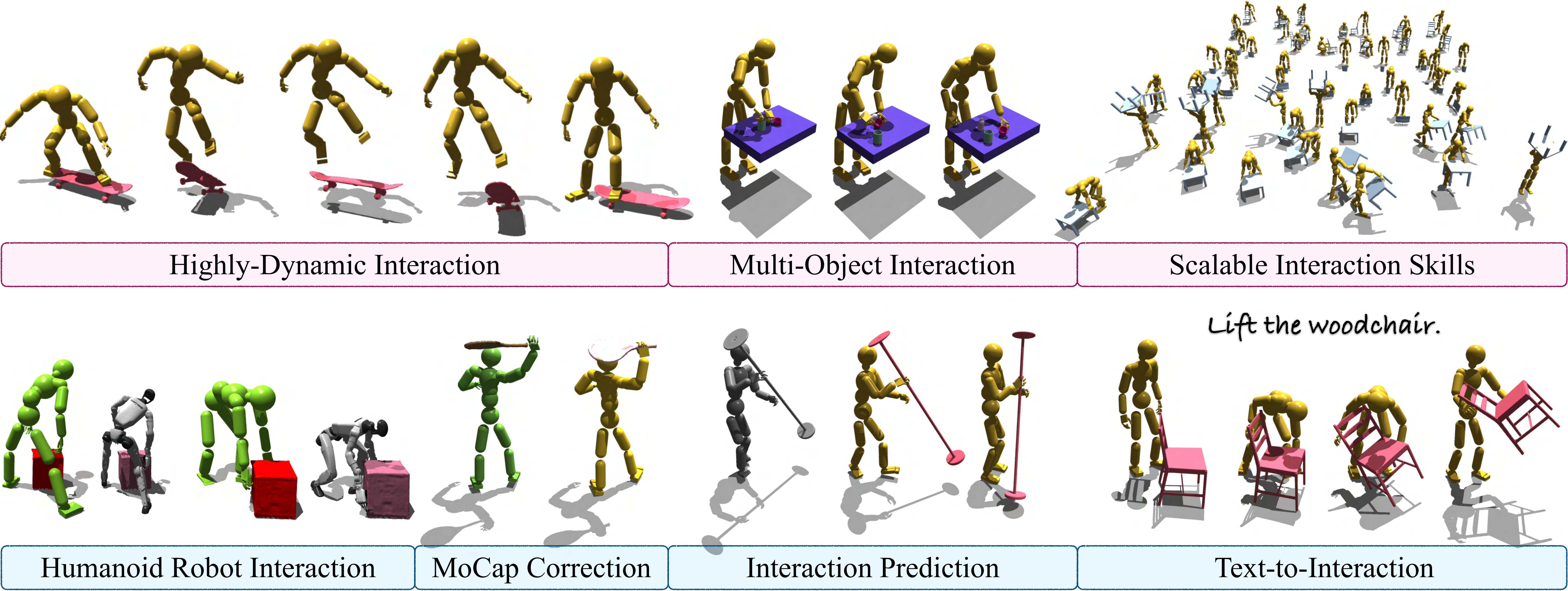}
\captionof{figure}{InterMimic enables physically simulated humans to perform interactions with dynamic and diverse objects. It supports highly-dynamic, multi-object interactions and scalable skill learning (\textbf{Top}), making it adaptable for versatile downstream applications (\textbf{Bottom}): it can translate whole-body loco-manipulation skills to a humanoid robot~\cite{unitreeg1,inspire}, perfect interaction MoCap data, and bridge kinematic generation, \eg, predicting future interactions from past (InterDiff~\cite{xu2023interdiff}) or generating interactions given text prompts (InterDreamer~\cite{xu2024interdreamer}).
\label{fig:teaser}}
\vspace{0.8em}
\end{strip}
\etocdepthtag.toc{mtchapter}
\begin{abstract}
Achieving realistic simulations of humans interacting with a wide range of objects has long been a fundamental goal. Extending physics-based motion imitation to complex human-object interactions (HOIs) is challenging due to intricate human-object coupling, variability in object geometries, and artifacts in motion capture data, such as inaccurate contacts and limited hand detail.
We introduce InterMimic, a framework that enables a single policy to robustly learn from hours of imperfect MoCap data covering diverse full-body interactions with dynamic and varied objects. Our key insight is to employ a curriculum strategy -- perfect first, then scale up. We first train subject-specific teacher policies to mimic, retarget, and refine motion capture data.
Next, we distill these teachers into a student policy, with the teachers acting as online experts providing direct supervision, as well as high-quality references. Notably, we incorporate RL fine-tuning on the student policy to surpass mere demonstration replication and achieve higher-quality solutions.
Our experiments demonstrate that InterMimic produces realistic and diverse interactions across multiple HOI datasets. The learned policy generalizes in a zero-shot manner and seamlessly integrates with kinematic generators, elevating the framework from mere imitation to generative modeling of complex human-object interactions.
\end{abstract}    
\section{Introduction}

Animating human-object interactions is a challenging and time-consuming task even for skilled animators. It requires a deep understanding of physics and meticulous attention to detail to create natural and convincing interactions. While Motion Capture (MoCap) data provides references, animators often need to correct contact errors caused by sensor limitations and occlusions between humans and objects. However, this process remains unscalable, as refining a single motion demands a delicate balance between preserving the captured data and ensuring its physical plausibility.

Physics-based human motion imitation~\cite{lee2010data,peng2018deepmimic} offers an alternative approach to improving motion fidelity, by training control policies to mimic reference MoCap data within a physics simulator. However, scaling up human-object interaction imitation presents significant challenges: (\textbf{i}) \textcolor{Sepia}{\textit{MoCap Imperfection}}: Contact artifacts are common, causing expected contacts to fluctuate instead of maintaining consistent zero distance, often due to MoCap limitations or missing hand capture~\cite{bhatnagar22behave,li2023object}. Accurately imitating MoCap kinematics can result in unrealistic dynamics in simulation. Moreover, HOI datasets often include diverse human shapes, requiring motion retargeting to adapt movements across different human models while preserving interaction dynamics. This retargeting process is imperfect and can introduce new contact artifacts or exacerbate existing ones.
(\textbf{ii}) \textcolor{Violet}{\textit{Scaling-up}}: Although large-scale motion imitation has been explored in previous works~\cite{yuan2020residual, luo2023perpetual, tessler2024maskedmimic, won2020scalable}, it remains largely underexplored for whole-body interactions involving dynamic and diverse objects.

In this paper, we aim to utilize rich yet imperfect motion capture interaction datasets to train a control policy capable of learning diverse motor skills while enhancing the plausibility of these actions by correcting errors, such as inaccurate hand motions and faulty contacts. Our approach is grounded on the key insight of tackling the challenges of \textcolor{Sepia}{\textit{skill perfection}} and \textcolor{Violet}{\textit{skill integration}} progressively. We implement a curriculum-based teacher-student distillation framework, where multiple teacher policies focus on imitating and refining small subsets of interactions, and a student policy integrates these skills from the teachers.

Instead of relying on curated data that covers a limited range of actions~\cite{luo2024grasping,braun2023physically}, we employ multiple teacher policies trained on a diverse set of imperfect interaction data and address two key challenges: \textit{retargeting} and \textit{recovering}.
First, we unify all training policies to a canonical human model, by embedding HOI retargeting directly into the imitation. This is achieved by reframing the policy learning to optimize both imitation and retargeting objectives.
Second, our teacher policies refine interaction motion through learning from it, as accurate contact dynamics enforced by a physics simulator inherently correct inaccuracies in the reference kinematics. To support this, we introduce tailored contact-guided reward and optimize trajectory collection, enabling effective skill imitation despite MoCap errors.

Introducing teacher policies offers several key benefits. By leveraging teacher rollouts, we effectively distill raw MoCap data into refined HOI references with a unified embodiment and enhanced physical fidelity. These refined references guide the subsequent student policy training, reducing the negative impact of errors in the original MoCap data.
A major hurdle in scaling motion imitation is the sample inefficiency of Reinforcement Learning (RL), which can lead to prohibitively long training times. Our teacher-student approach mitigates this through a \textit{space-time trade-off}: multiple teacher policies are trained in parallel on smaller, more manageable data subsets, and their expertise is then \textit{distilled} into a single student policy. We begin with demonstration-based distillation to bootstrap PPO~\cite{schulman2017proximal} updates, reducing reliance on pure trial and error and enabling more effective scaling. As training progresses, the student gradually shifts from heavy demonstration guidance to increased RL updates, ultimately surpassing simple demonstration memorization. This mirrors alignment strategies in Large Language Models (LLMs), where demonstration-based pretraining is refined through RL fine-tuning~\cite{ouyang2022training}.

To summarize, our contributions are as follows:  
(\textbf{i}) We introduce \textit{InterMimic}, which, to the best of our knowledge, is the \textit{first} framework designed to train physically simulated humans to develop \textit{a wide range of whole-body} motor skills for interacting with \textit{diverse} and \textit{dynamic} objects, extending beyond traditional grasping tasks.  
(\textbf{ii}) We develop a teacher-student training strategy, where teacher policies provide a unified solution to address the challenges of retargeting and refining in HOI imitation. The student distillation introduces a scalable solution by leveraging a space-time trade-off.  
(\textbf{iii}) We demonstrate that our unified framework, \textit{InterMimic}, as illustrated in Figure~\ref{fig:teaser}, effectively handles versatile physics-based interaction animation, recovering motions with realistic and physically plausible details. Notably, by combining kinematic generators with \textit{InterMimic}, we enable a physics-based agent to achieve tasks such as interaction prediction and text-to-interaction generation.

\section{Related Work}
Significant progress has been made in physics-based human interaction animation and control, with advancements in areas such as human-human interactions~\cite{liu2024physreaction,wang2023intercontrol}, hand-object interactions~\cite{park2025learningtransferhumanhand, yang2022learning, wang2024furelise, xu2024synchronize}, human interactions with static scenes~\cite{chao2021learning, pan2023synthesizing, xiao2024unified, tessler2024maskedmimic, lee2023locomotion}, and real-world humanoid control for object manipulation~\cite{gu2025humanoidlocomotionmanipulationcurrent, sferrazza2024humanoidbench, he2024omnih2o, ze2024generalizable, jiang2024dexmimicgen, he2024learning, ben2024homie,li2024okami,dao2024sim,liu2024opt2skill}. Among these areas, we are the first to achieve universal whole-body loco-manipulation simulation with diverse dynamic objects, beyond pick-and-place and grasping actions -- a novel achievement in animation and an idealized reference for real-world humanoid control.
Below, we elaborate on recent studies on \textit{whole-body} interaction animation, particularly involving \textit{dynamic objects}.

\subsection{Kinematic Interaction Animation}
Generating human interactions has been a long-standing topic in animation and computer graphics~\cite{lee2006motion, gleicher1997motion}. Significant advances in character animation have emerged with the advent of deep learning, \eg, including phase-function-based methods~\cite{holden2017phase} that enable object interactions like carrying a box~\cite{starke2019neural} or playing basketball~\cite{starke2020local}. This is extended to more diverse but static objects approaching~\cite{zhang2022couch, wu2022saga, taheri2022goal, kulkarni2023nifty}.
Subsequent efforts~\cite{ghosh2022imos, razali2023action, li2023controllable, wu2024human, li2023task, jiang2024scaling, jiang2024autonomous, lu2024choice} integrate object motion into interactions but remain constrained by assuming that interactions occur primarily through the hands. To address this, recent developments~\cite{corona2020context, 9714029, xu2023interdiff, peng2023hoi, diller2023cg, wu2024thor, daiya2024collage, song2024hoianimator, david, xu2024interdreamer, he2024syncdiff} introduce interactions in a fashion of whole-body loco-manipulation that engages multiple body parts in contact. However, these methods often suffer from physical inaccuracies, such as floating contacts and penetrations, while they generate only body motion without considering hand dexterity.
In this work, we address physical inaccuracies by refining imperfect kinematic generation through physics simulation, with InterDiff~\cite{xu2023interdiff} and HOI-Diff~\cite{peng2023hoi} serving as motion planning for loco-manipulation that bridges high-level decision-making (\eg, text instruction) with low-level execution.

\subsection{Physics-based Interaction Animation}
Physics-based methods generate motion through motor control policies within a physics simulator, \eg, achieved via deep reinforcement learning to track reference motions~\cite{peng2018deepmimic}. These policies are directly applicable for executing simple interactions, such as punching or striking an object~\cite{peng2022ase, tessler2023calm, cui2024anyskill, tevet2024closd}. To achieve more complex interactions, early studies focus on specific scenarios, including notable sports-related~\cite{luo2024smplolympics} examples such as basketball~\cite{wang2023physhoi}, skating~\cite{liu2017learning}, soccer~\cite{xie2022learning}, tennis~\cite{zhang2023learning}, table tennis~\cite{wang2024strategy}, and more proposed in~\cite{bae2023pmp}.
Research also demonstrates flexibility in more general but simpler box carrying tasks~\cite{zhang2023simulation,pan2025tokenhsi,wang2024sims}. These advancements are achieved through the integration of multiple control policies~\cite{merel2020catch}, the use of adversarial motion priors~\cite{hassan2023synthesizing, peng2021amp,gao2024coohoi}, and imitating diverse kinematic generations~\cite{xie2023hierarchical,wu2024human}.
However, these methods train their policies in a \textit{non-scalable} manner, with each policy handling only specific object types or actions. In pursuit of a single, scalable policy to enable multiple interaction skills, existing methods either rely on fixed interaction patterns, such as approaching and grasping objects~\cite{braun2023physically,luo2024grasping}, or extend single-object skills, \eg, interactions involving a basketball~\cite{wang2024skillmimic}. Additionally, they mostly depend on highly curated data from the GRAB dataset~\cite{taheri2020grab}, which, despite its high quality, primarily features low-dynamic full-body motion and only small-sized objects. More recent datasets~\cite{bhatnagar22behave, jiang2022chairs, huang2022intercap, zhang2023neuraldome, li2023object, zhao2023im, kim2024parahome, wu2024himo, zhang2024core4d, xie2024intertrack, zhang2024hoi, zhang2024force, liu2024mimicking, liu2024taco, xu2025interact} offer richer full-body interactions with diverse objects but contain noticeable artifacts that challenge existing motion imitation approaches. We process data from OMOMO~\cite{li2023object}, BEHAVE~\cite{bhatnagar22behave}, HODome~\cite{zhang2023neuraldome}, and IMHD~\cite{zhao2023im} collected in the InterAct~\cite{xu2025interact} dataset, and the multi-object dataset HIMO~\cite{wu2024himo}, highlighting InterMimic's \textit{scalability} to diverse interactions and its \textit{robustness} against MoCap artifacts.
\begin{figure}
    \centering
    \includegraphics[width=\columnwidth]{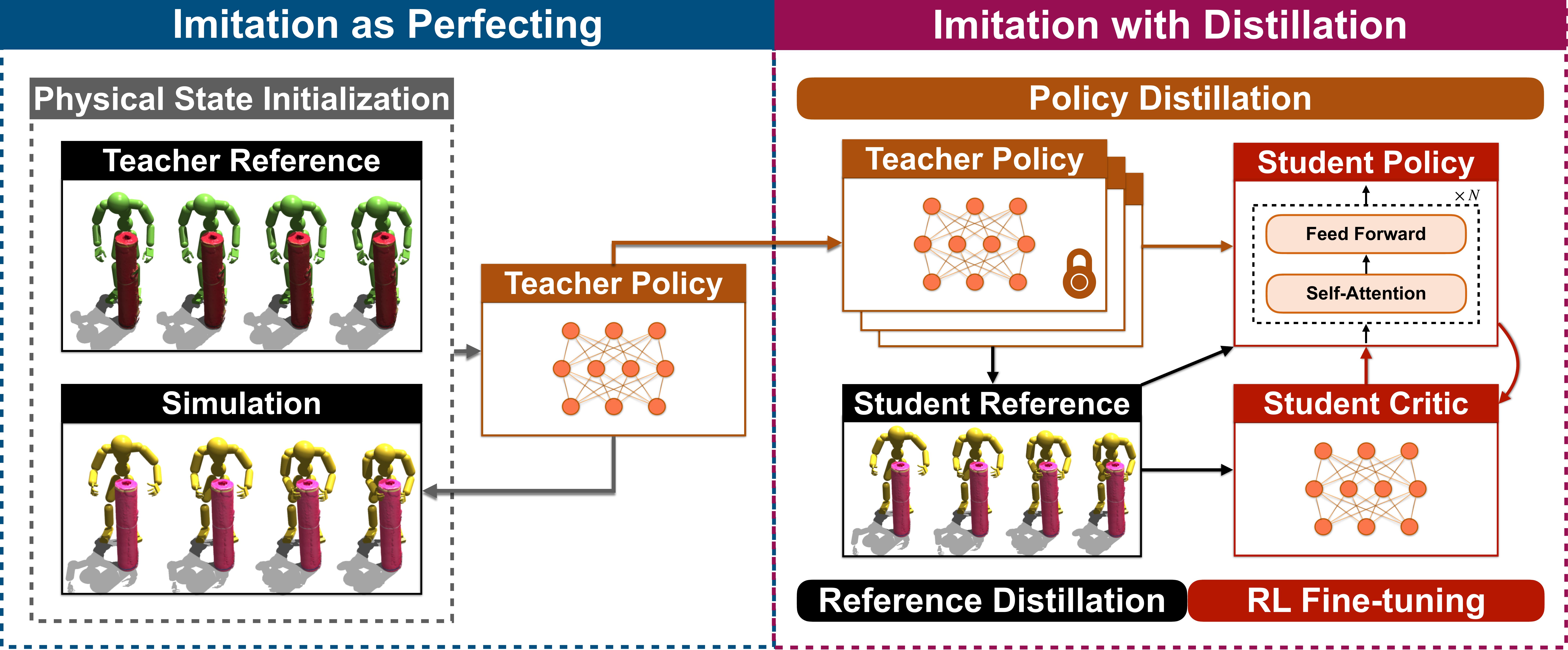}
    \caption{Our two-stage pipeline: (\textbf{i}) training each teacher policy (MLP) on a small data subset with initialization corrected via Physical State Initialization (PSI), and (\textbf{ii}) freezing the teacher policies to provide refined references for training a student policy (Transformer). The student leverages teacher supervision for effective scaling and is fine-tuned through RL.
    }
    \label{fig:method}
\end{figure}
\section{Methodology}
\noindent\textbf{Task Formulation.}
The goal of human-object interaction (HOI) imitation is to learn a policy \(\pi\) that produces simulated human-object motion \(\{\boldsymbol{q}_t\}_{t=1}^T\) closely matching a ground-truth reference \(\{\hat{\boldsymbol{q}}_t\}_{t=1}^T\) derived from large-scale MoCap data. Given the geometries of the human and objects, the policy should also compensate for missing or inaccurate details in the dataset.  
Each pose \(\boldsymbol{q}_t\) has two components: the human pose \(\boldsymbol{q}^h_t\) and the object pose \(\boldsymbol{q}^o_t\). The human pose is defined as \(\boldsymbol{q}^h_t = \{\boldsymbol{\theta}^h_t, \boldsymbol{p}^h_t\}\), where \(\boldsymbol{\theta}^h_t \in \mathbb{R}^{52 \times 3}\) represents the joint rotations, and \(\boldsymbol{p}^h_t \in \mathbb{R}^{52 \times 3}\) specifies the joint positions. Specifically, our human model includes 30 hand joints and 22 joints for the rest of the body, with one root joint’s rotation and position specified in global coordinates, whereas the rotations and positions of all other joints are defined relative to their respective parent joints' coordinate systems.
The object pose \(\boldsymbol{q}^o_t\) is represented as \(\{\boldsymbol{\theta}^o_t, \boldsymbol{p}^o_t\}\), where \(\boldsymbol{\theta}^o_t \in \mathbb{R}^{3}\) denotes the object’s orientation and \(\boldsymbol{p}^o_t \in \mathbb{R}^{3}\) the position.
All simulation states have corresponding ground-truth values, denoted by the hat symbol. For instance, the reference object rotation is \(\{\hat{\boldsymbol{\theta}}^o_t\}_{t=1}^T\). The environmental setup for the simulation is detailed in Sec.~\ref{sec:phys_para}.

\noindent\textbf{Overview.}
We formulate interaction imitation as a Markov Decision Process (MDP), defined by states, actions, simulator-provided transition dynamics, and a reward function.  
Figure~\ref{fig:method} illustrates our two-stage framework: (i) training teacher policies \(\pi^{(T)}\) on small skill subsets, and (ii) distilling these teachers into a scalable student policy \(\pi^{(S)}\) for large-scale skill learning.  
In Sec.~\ref{sec:state}, we define the states \(\boldsymbol{s}_t\) and actions \(\boldsymbol{a}_t\), applicable to both teacher \(\pi^{(T)}\) and student \(\pi^{(S)}\) policies.  
In Sec.~\ref{sec:teacher}, we describe how teacher policies are trained via RL, focusing on reward designs that facilitate retargeting, as well as techniques that mitigate the impact of imperfections in the reference data.  
Sec.~\ref{sec:student} details the subsequent distillation of teachers into a scalable student policy, leveraging both RL and learning from demonstration.

\subsection{Policy Representation} \label{sec:state}
\noindent\textbf{State.}
The state \(\boldsymbol{s}_t\), which serves as input to the policy, comprises two components \(\boldsymbol{s}_t = \{\boldsymbol{s}_t^s, \boldsymbol{s}_t^g\}\). The first part, \(\boldsymbol{s}_t^s\), contains human proprioception and object observations, expressed as,
 $\{\{\boldsymbol{\theta}_t^h, \boldsymbol p_t^h, \boldsymbol{\omega}_t^h, \boldsymbol v_t^h\}, \{\boldsymbol{\theta}_t^o, \boldsymbol p_t^o, \boldsymbol{\omega}_t^o, \boldsymbol v_t^o\}, \{\boldsymbol{d}_t, \boldsymbol{c}_t\}\},$
where $\{\boldsymbol{\theta}_t^h, \boldsymbol p_t^h, \boldsymbol{\omega}_t^h, \boldsymbol v_t^h\}$ represent the rotation, position, angular velocity, and velocity of all joints, respectively, while $\{{\boldsymbol \theta}_t^o,  \boldsymbol p_t^o, {\boldsymbol \omega}_t^o,  \boldsymbol v_t^o\}$ represent the orientation, location, velocity, and angular velocity of the object, respectively. Motivated by~\cite{christen2022d}, we include object geometry and whole-body haptic sensing from two elements: (\textbf{i}) $\boldsymbol{d}_t$, vectors from human joints to their nearest points on each object surface; and (\textbf{ii}) $\boldsymbol{c}_t$, contact markers indicating whether the human’s rigid body parts experience applied forces; this serves as simplified tactile or force sensing -- an important multi-modal input in robot manipulation tasks~\cite{dahiya2009tactile}. The goal state $\boldsymbol s_{t}^g = \{\boldsymbol s_{t, t+k}^g\}_{k \in K}$ integrates reference poses from the ground truth motion, where $\boldsymbol s_{t, t+k}^g$ is defined as,
\begin{equation}\label{eq:state}
\begin{aligned} 
 \{\{\hat{\boldsymbol{\theta}}_{t+k}^h \ominus \boldsymbol{\theta}_t^h, \hat{\boldsymbol p}_{t+k}^h - \boldsymbol p_t^h\}, \{\hat{{\boldsymbol \theta}}_{t+k}^o \ominus {\boldsymbol \theta}_t^o, \hat{ \boldsymbol p}_{t+k}^o -  \boldsymbol p_t^o\}, \\\{\hat{\boldsymbol{d}}_{t+k} - \boldsymbol{d}_t, \hat{\boldsymbol{c}}_{t+k} - \boldsymbol{c}_t\}, \{\hat{\boldsymbol{\theta}}_{t+k}^h, \hat{\boldsymbol p}_{t+k}^h, \hat{{\boldsymbol \theta}}_{t+k}^o, \hat{\boldsymbol p}_{t+k}^o \}\},
\end{aligned} 
\end{equation}
where $\hat{\boldsymbol{\theta}}_{t+k}^h, \hat{\boldsymbol p}_{t+k}^h, \hat{\boldsymbol{d}}_{t+k}, \hat{\boldsymbol{c}}_{t+k}$ represent the reference information at time step $t+k$, $\ominus$ denotes the calculation of rotation difference.
All continuous elements of $\boldsymbol{s}_t$ are normalized relative to the current direction of view of the human and the position of the root~\cite{peng2018deepmimic}. 

Given that most MoCap data does not provide reference contact or tactile information, we extract reference contact markers \(\hat{\boldsymbol{c}}_{t+k}\) by inferring dynamic information, beyond relying solely on inaccurate contact distances, specifically by analyzing the object's acceleration to detect human-induced forces. 
To accommodate the variability in contact distances observed in reference motion, we discretize reference contact markers using varying distance thresholds, as illustrated in Fig.~\ref{fig:contact_label}(\textbf{i}). The neutral areas serve as buffer zones, avoiding the penalization or enforcement of strict contact.
See Sec.~\ref{sec:repre_supp} of supplementary for details.

\noindent\textbf{Action.}
Our human model has 51 actuated joints, defining an action space of $\boldsymbol{a}_t \in \mathbb{R}^{51\times3}$. These actions are specified as joint PD targets using the exponential map and are converted into torques applied to each of the human joints.

\subsection{Imitation as Perfecting}\label{sec:teacher}
The teacher policy \(\pi^{(\text{T})}\) is trained via RL to maximize the expected discounted reward by comparing simulated states against potentially erroneous reference states. The training involves: (\textbf{i}) trajectory collection, where we explain how trajectories are initialized and terminated. (\textbf{ii}) policy updating, where collected trajectories and their associated rewards are used to refine the policy. In this section, we elaborate on our reward design and how we optimize trajectory collection to mitigate the impact of reference inaccuracies.

\noindent\textbf{Imitation as Retargeting.}
We tailor teacher policies to each human subject, while all policies share the same base human model. This serves the retargeting purpose by converting HOIs from different human shapes into a unified base shape. Although motion imitation does not necessarily require a unified human model~\cite{won2019learning,luo2023perpetual}, our approach offers two benefits:
(i) It enhances integration with kinematic generation methods, which generally perform better on a single, unified shape~\cite{guo2022generating}.
(ii) It demonstrates possible integration with real-world humanoid deployment, which requires retargeting to a consistent physical embodiment. In Figure~\ref{fig:teaser}, our method translates MoCap data into motor skills on a Unitree G1~\cite{unitreeg1} with two Inspire hands~\cite{inspire}, all without external retargeting in complex contact-rich scenarios. See Sec.~\ref{sec:training} of the supplementary for additional details.

Human~\cite{villegas2021contact} or HOI~\cite{kim2016retargeting} retargeting can be formulated as an optimization problem. Inverse Kinematics (IK) methods, such as those based on quadratic programming~\cite{kraft1994algorithm}, demonstrate effectiveness in simplified scenarios but remain underexplored for motions featuring intricate object interactions. RL, by contrast, solves the optimization by maximizing an expected cumulative reward, prompting us to investigate whether RL-driven HOI imitation can be used for HOI retargeting. This extends existing physics-based retargeting approaches, which either omit object interactions~\cite{reda2023physics} or are non-scalable with a single reference~\cite{zhang2023simulation}.

While the kinematics should differ due to the embodiment gap, we argue that the \textit{dynamics} between human and object should remain \textit{invariant}. Thus, we define rewards to include an embodiment-aware component that loosely aligns the simulated kinematics with the reference interaction, and an embodiment-agnostic reward component that encourages dynamics to be close to the reference.

\noindent\textbf{Embodiment-Aware Reward.} When the human and object are far apart, retargeting should prioritize capturing rotational motion, whereas when they are close, accurate position tracking becomes crucial for achieving contact. To reflect this, we define the weights \( \boldsymbol{w}_d \) that are inversely proportional to the distances between joints and the object~\cite{zhang2023simulation}. The reward thus includes cost functions for joint position \( E_p^h = \langle \boldsymbol\Delta^h_{p}, \boldsymbol{w}_d \rangle\), rotation \( E_{\theta}^h = \langle \boldsymbol\Delta^h_{\theta}, \boldsymbol 1 - \boldsymbol{w}_d \rangle\), and interaction tracking \( E_d = \langle \boldsymbol\Delta_{d}, \boldsymbol{w}_d \rangle \), where \(\langle \cdot, \cdot \rangle\) is the inner product, $\boldsymbol\Delta^h_{p}[i]=\|\hat{\boldsymbol{p}}^h[i] - \boldsymbol{p}^h[i]\|$, $\boldsymbol\Delta^h_{\theta}[i]=\|\hat{\boldsymbol{\theta}}^h[i] \ominus \boldsymbol{\theta}^h[i]\|$, and $\boldsymbol\Delta_{d}[i]=\|\hat{\boldsymbol{d}}[i] - \boldsymbol{d}[i]\|$ represent the displacement for the variables defined in Sec.~\ref{sec:state} with timestep $t$ omitted. The formulation of \( \boldsymbol{w}_d \) is provided in Sec.~\ref{sec:reward} of supplementary. The reward to be maximized can be formulated as $\exp(-\lambda E)$ for each cost function $E$ with a specific hyperparameter $\lambda$. Details can be found in Sec.~\ref{sec:reward}.

\noindent\textbf{Embodiment-Agnostic Reward.} The reward includes components for object tracking and contact tracking. The object tracking cost is defined for position $E^o_p = \|\hat{\boldsymbol{p}}^{o} -  \boldsymbol p^{o}\|$ and rotation $E^o_{\theta} = \|\hat{\boldsymbol\theta}^{o} - {\boldsymbol\theta}^{o}\|$, with all values normalized to the human's current position and direction.
\begin{figure}
    \centering
    \includegraphics[width=\columnwidth]{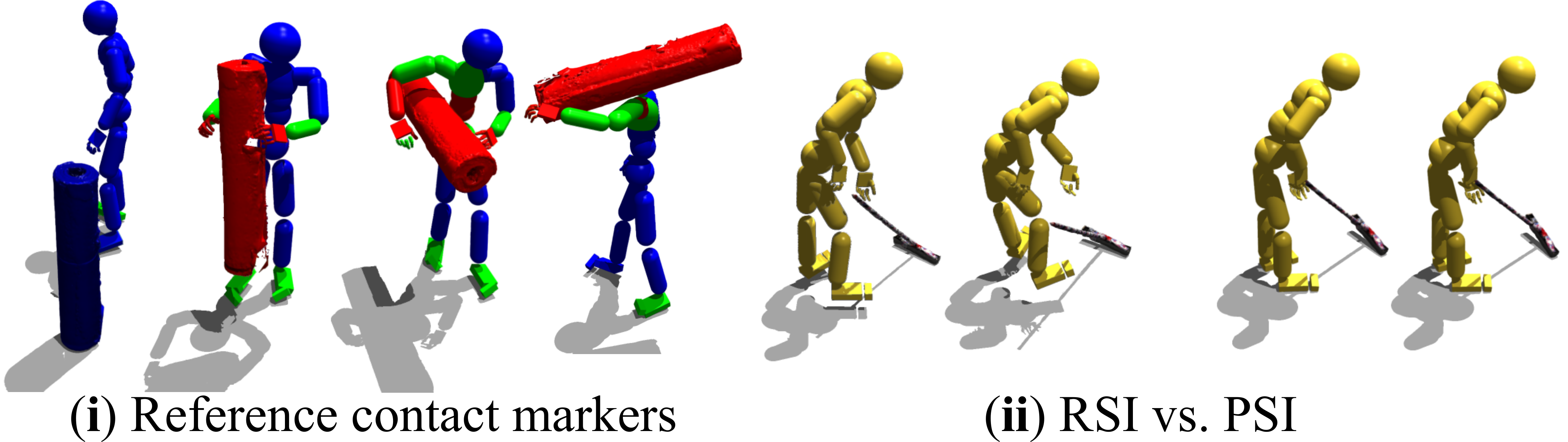}
    \caption{(\textbf{i}) Visualization of reference contact markers that accommodate varied contact distances: \textcolor{red}{red} to promote contact, \textcolor{green}{green} for neutral areas where contact is neither promoted nor penalized, and \textcolor{blue}{blue} to penalize contact. (\textbf{ii}) Initializing the rollout with reference (RSI) or reference corrected via simulation (PSI).
    }
    \label{fig:contact_label}
\end{figure}
The contact tracking reward comprises two cost functions: body contact promotion \( E^c_b \) and penalty \( E^c_p \), both aligning the simulated contact \( \boldsymbol{c} \) with reference markers \( \hat{\boldsymbol{c}} \), as shown in Figure~\ref{fig:contact_label}. We define three contact levels -- promotion, penalty, and neutral -- to accommodate potential inaccuracies in reference contact distances. The detailed formulation can be found in Sec.~\ref{sec:reward} of the supplementary.
Since the physics engine does not differentiate between object, ground, and self-contact, we adopt two strategies:
\textbf{(i)} we model foot-ground contact promotion and penalty. This ensures proper foot lifting during cyclic walking and mitigates foot hobbling.
\textbf{(ii)} We allow self-collision to avoid self-contact promotion but to promote object interaction. This poses minimal risk as the policy is guided by MoCap reference, which, although lacking perfect contact accuracy, rarely shows self-penetration. For humanoid robots with embodiments that differ from the MoCap reference and require real-world applicability, we disable self-collision, as discussed in Sec.~\ref{sec:training}.

We introduce energy consumption rewards~\cite{yu2018learning} to penalize large human or object jitters, with a proposed contact energy penalizing abrupt contact to promote compliant interactions. See Sec.~\ref{sec:reward} of supplementary for more details.

\noindent\textbf{Hand Interaction Discovery.}
We use data with average or flattened hand poses~\cite{bhatnagar22behave,li2023object}, which makes accurate object manipulation imitation challenging. To address this, we activate a reference contact marker for any hand part when a \textit{fingertip} or \textit{palm} is near an object. Given tasks that do not demand high dexterity, employing a contact-promoting reward with this marker enables policies to develop effective hand interaction strategies, leveraging the exploratory nature of RL. Additionally, we constrain the range of motion (RoM) of the hands to ensure natural movement. See Sec.~\ref{sec:reward} and Sec.~\ref{sec:phys_para} of the supplementary for further details.

\noindent\textbf{Policy Learning.}
Following~\cite{peng2018deepmimic}, the control policy $\pi$ is trained using PPO~\cite{schulman2017proximal} with the policy gradient $L(\psi) = \mathbb{E}_t [ \min(r_t(\psi) {A}_t, \text{clip}(r_t(\psi), 1-\epsilon, 1+\epsilon) {A}_t).$ $\psi$ are the parameters of $\pi$ and \( r_t(\psi) \) quantifies the difference in action likelihoods between updated and old policies. \( \epsilon \) is a small constant, and \({A}_t \) is the advantage estimate given by the generalized advantage estimator GAE($\lambda$)~\cite{schulman2015high}.

\noindent\textbf{Physical State Initialization.}
Learning later-phase motion can be essential for policies to achieve high rewards during earlier phases, compared to incrementally learning from the starting phase. 
Thus, Reference State Initialization (RSI)~\cite{peng2018deepmimic} sets the current pose ${\boldsymbol{q}}_t$ to a reference pose $\hat{\boldsymbol{q}}_t$ at a random timestep $t$, for initializing the rollout.
However, initializing with the imperfect reference can introduce \textit{critical artifacts}, such as contact floating or incorrect hand motion, leading to unrecoverable failures, \eg, object falling, as depicted in Figure~\ref{fig:contact_label}(ii). These issues render many initializations ineffective, limiting training on certain interaction phases since successful rollouts may not reach them before the maximum length. The problem is exacerbated by the use of prioritized sampling~\cite{won2019learning,tessler2024maskedmimic}, which favors high-failure-rate initializations.

To address the need for higher-quality reference initialization, we propose \textit{Physical State Initialization} (PSI). As illustrated in Figure~\ref{fig:method}, PSI begins by creating an initialization buffer that stores reference states from MoCap and simulation states from prior rollouts. For each new rollout, an initial state is randomly selected from this buffer, which increases the likelihood of starting from advantageous positions. Once a rollout is completed, trajectories are evaluated based on their expected discounted rewards; those above a certain threshold are added to the buffer using a first-in-first-out (FIFO) strategy, while older or lower-quality trajectories are discarded. This selective reintroduction of high-value states for initialization helps maintain stable policy updates. We apply PSI in a sparse manner to ensure training efficiency. As shown in Figure~\ref{fig:contact_label}(ii), PSI can collect trajectories for policy update that RSI does not effectively utilize. Further details are provided in Sec.~\ref{sec:psi_supp} of the supplementary.

\noindent\textbf{Interaction Early Termination.}
Early Termination (ET)~\cite{peng2018deepmimic} is commonly used in motion imitation, ending an episode when a body part makes unplanned ground contact or when the character deviates significantly from the reference~\cite{luo2023perpetual}, thus stopping the policy from overvaluing invalid transitions. However, additional conditions should be considered for human-object interactions. We propose \textit{Interaction Early Termination} (IET), which supplements ET with three extra checks:
(\textbf{i}) Object points deviate from their references by more than 0.5\,m on average.
(\textbf{ii}) Weighted average distances between the character’s joints and the object surface exceed 0.5\,m from the reference.
(\textbf{iii}) Any required body-object contact is lost for over 10 consecutive frames.
Full conditions are detailed in Sec.~\ref{sec:psi_supp} of the supplementary.

\subsection{Imitation with Distillation}\label{sec:student}
As shown in Figure~\ref{fig:method}, after training the teacher policies on data from each subject (Sec.~\ref{sec:teacher}), we aggregate them to train a student policy $\pi^{(\text{S})}$ to master all skills. As outlined in Algorithm~\ref{algo:dis}, the combined teacher policies, denoted by $\pi^{(\text{T})}$ for brevity, serves dual roles by providing state-action trajectories $(\boldsymbol s^{(\text{T})}, \boldsymbol a^{(\text{T})})$: (\textbf{i}) the state $\boldsymbol s^{(\text{T})}$ for reference distillation, and (\textbf{ii}) the action $\boldsymbol a^{(\text{T})}$ for policy distillation.

\noindent\textbf{Reference Distillation.}
Noisy MoCap data can hinder policy learning, especially at larger scales. In contrast, teacher 
policies trained on smaller-scale data effectively address these issues by correcting contact artifacts, refining hand placements, and recovering missing details (see Figures~\ref{fig:teaser} and \ref{fig:obj_rot}). To fully leverage teacher policies, we use their rollouts as references for defining the student policy’s goal state and reward functions, distinguishing our approach from distillation based on only action output.

\noindent\textbf{Policy Distillation.} We apply distillation on action outputs, which we view as crucial for scaling policies to large datasets. In essence, we trade space for time: teacher policies are trained in parallel on smaller data subsets, allowing the student policy to scale through distillation.
Following Algorithm~\ref{algo:dis}, we begin with Behavior Cloning (BC)~\cite{juravsky2024superpadl} and use RL fine-tuning to go beyond demonstration memorization, an approach common in LLM alignment~\cite{ouyang2022training}. We integrate BC into online policy updates with a staged schedule: we start with DAgger~\cite{ross2011reduction} and gradually transition to PPO. Throughout, the critic is continuously trained with the reward from Sec.~\ref{sec:teacher}. This RL fine-tuning phase is crucial as teacher policies may behave differently when performing similar skills, and simple BC can lead to suboptimal ``averaging'' behavior, where RL fine-tuning helps the student converge on optimal solutions.

\subsection{Architecture}
We set the keyframe indices \(K\) (Sec.~\ref{sec:state}, Eq.~\ref{eq:state}) to \(\{1, 16\}\) for the teacher policies and \(\{1, 2, 4, 16\}\) for the student policy. The broader observation window for the student policy helps it better distinguish different skills with larger-scale data. Teacher policies employ MLPs, common in physics-based animation~\cite{peng2018deepmimic}, while the student policy handles higher-dimensional observations, for which MLPs are less effective. Thus, we use a transformer~\cite{transformer} architecture for sequential modeling~\cite{tessler2024maskedmimic}, as shown in Figure~\ref{fig:method}.

\begin{algorithm}
    \caption{Distillation with RL Fine-tuning}
    \label{algo:dis}
    \begin{algorithmic}[1]
    \State \textbf{Input}: A composite policy $\pi^{(\text{T})}$ integrated from individual teacher policies, student policy parameters $\boldsymbol{\psi}$, student value function parameters $\boldsymbol{\phi}$, schedule hyperparameter $\beta$ for DAgger, horizon length $H$ for PPO
    \For{$t = 0, 1, 2, \ldots$ } 
        \For{$h$ from $1$ to $H$}
            \State Sample a variable $u \sim \text{Uniform}(0, 1)$
            \State Collect $\boldsymbol s^{(\text{T})}, \boldsymbol a^{(\text{T})}$ from teacher $\pi^{(\text{T})}$
            \State Obtain the refined reference from $\boldsymbol s^{(\text{T})}$ to define $\boldsymbol{s}^{(\text{S})}$ and $r(\cdot)$, obtain $\boldsymbol a^{(\text{S})}$ from $\pi^{(\text{S})}_{\boldsymbol\phi}(\boldsymbol a^{(\text{S})}|\boldsymbol{s}^{(\text{S})})$.
            \If{$u \leq \frac{t}{\beta}$} \Comment{Use the teacher}
                \State Given $\boldsymbol s^{(\text{S})}$, execute $\boldsymbol a^{(\text{S})}$, observe $\boldsymbol s'^{(\text{S})}, r$
            \Else \Comment{Use the student}
                \State Given $\boldsymbol s^{(\text{S})}$, execute $\boldsymbol a^{(\text{T})}$, observe $\boldsymbol s'^{(\text{S})}, r$
            \EndIf
            \State Store the transition $(\boldsymbol s^{(\text{S})}, \boldsymbol s'^{(\text{S})}, \boldsymbol a^{(\text{S})}, \boldsymbol a^{(\text{T})}, r)$
        \EndFor
    \State Update $\boldsymbol\phi$ with TD($\lambda$)
    \State Compute PPO objective: $L(\psi)$
    \State Compute $J(\psi) = \|\boldsymbol a^{(\text{S})} - \boldsymbol a^{(\text{T})}\|$
    \State Compute the weight: $w = \min(\frac{t}{\beta}, 1)$
    \State Update $\psi$ by gradient: $\nabla_{\psi} (wL(\psi) + (1-w) J(\psi))$
    \EndFor
    \end{algorithmic}
\end{algorithm}
\section{Experiments}
We evaluate teacher policies on their ability to imitate imperfect HOI references, and assess the entire teacher-student framework for scalability to large-scale data and zero-shot generalization across various scenarios. Additional experiments are provided in Sec.~\ref{sec:add_exp} of supplementary.

\begin{figure}
    \centering
    \includegraphics[width=\columnwidth]{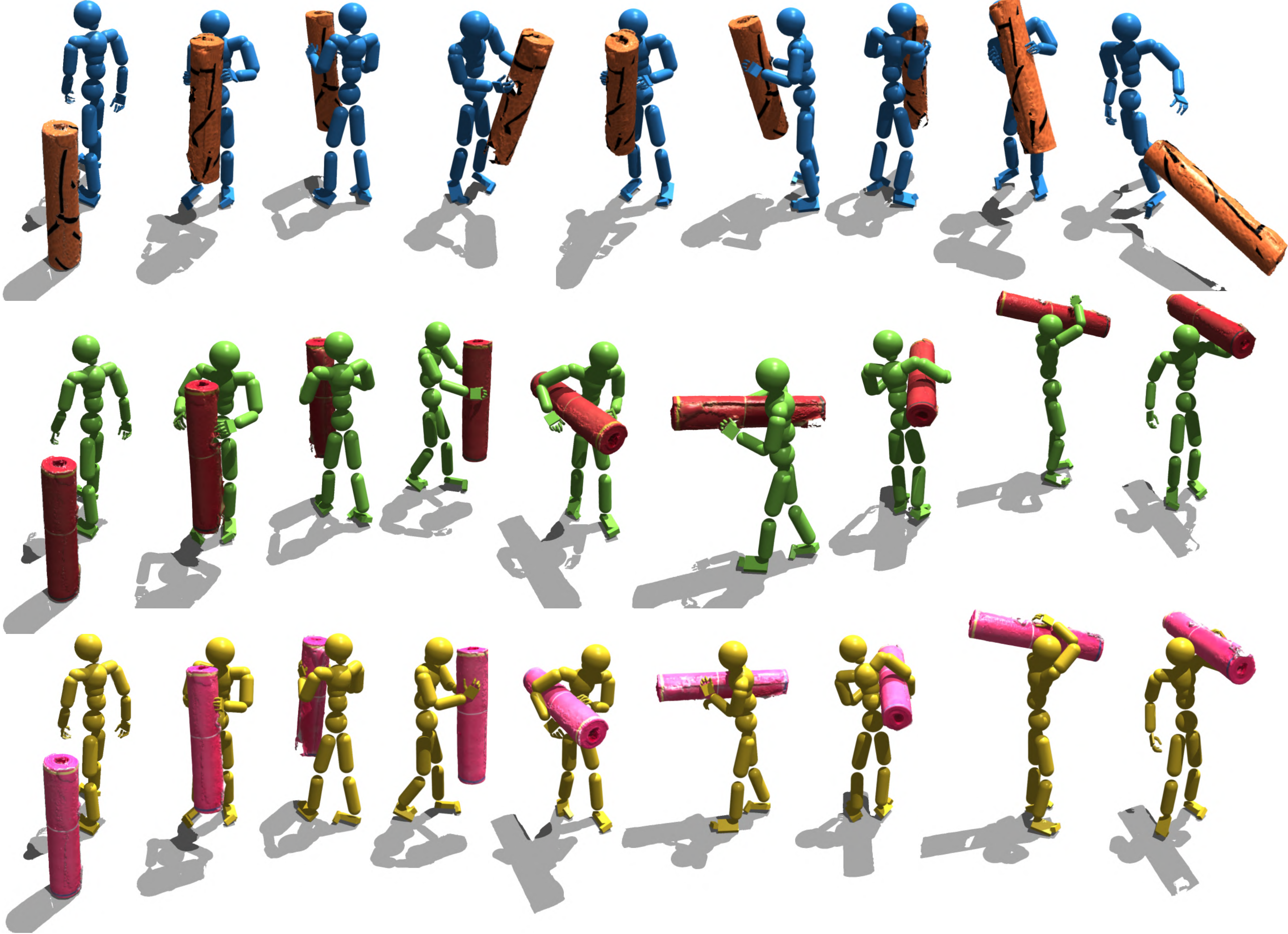}
    \caption{Qualitative comparison between PhysHOI~\cite{wang2023physhoi} (top), the reference motion (middle) from the BEHAVE~\cite{bhatnagar22behave} dataset, and the interaction refined by our teacher trained on it (bottom). InterMimic faithfully imitates the interactions involving multiple body parts while correcting errors in the original reference.}
    \label{fig:compare}
    \vspace{-0.5em}
\end{figure}

\noindent\textbf{Datasets.} 
We use the following datasets from InterAct~\cite{xu2025interact}: OMOMO~\cite{li2023object}, BEHAVE~\cite{bhatnagar22behave}, HODome~\cite{zhang2023neuraldome}, IMHD~\cite{zhao2023im}, and HIMO~\cite{wu2024himo}. OMOMO, containing 15 objects and approximately 10 hours of data, is our primary dataset for evaluating the full teacher-student distillation framework for its scale. We train 17 teacher policies, one per subject, with subject 14 reserved as the test set and the remaining data used for training the student policy. A small portion of data is discarded after teacher imitation due to severe MoCap errors that could not be corrected (see Sec.~\ref{sec:training} and Sec.~\ref{sec:discuss} of the supplementary).
Additional datasets are used to evaluate teacher policies in various MoCap scenarios with different error levels and interaction types. We focus on highly dynamic motions (Figure~\ref{fig:teaser}) and interactions involving multiple body parts (Figure~\ref{fig:compare}), while excluding scenarios such as carrying a bag with a strap, since the simulator~\cite{makoviychuk2021isaac} used lacks full support for soft bodies.

\noindent\textbf{Metrics.}
We use the following metrics: (\textbf{i}) \textit{Success Rate} is defined as the proportion of references that the policy successfully imitates at least once, averaged across all references, while (\textbf{ii}) \textit{Duration} is the time (in seconds) that the imitation is maintained without triggering the interaction early termination conditions introduced in Sec.~\ref{sec:teacher}. (\textbf{iii}) \textit{Human Tracking Error} ($E_h$) measures the per-joint position error (cm) between the simulated and reference human (excluding hand joints for BEHAVE~\cite{bhatnagar22behave} and OMOMO~\cite{li2023object} due to inaccuracy). (\textbf{iv}) \textit{Object Tracking Error} ($E_o$) measures the per-point position error (cm) between the simulated and reference object. Both errors are averaged over the duration of the imitation in the best-performing trial.

\noindent\textbf{Baselines.} 
To facilitate fair comparisons, we downgrade our method for teacher policy evaluation to imitate either a single MoCap clip (Figure~\ref{fig:compare}) or multiple clips with a single object (Table~\ref{table:comparison}), enabling direct comparison with PhysHOI~\cite{wang2023physhoi} and SkillMimic~\cite{wang2024skillmimic} (Sec.~\ref{sec:quan} and \ref{sec:qual}). Due to the lack of established baselines for large-scale HOI imitation, we adapt the following variants for comparison with our student policy (Sec.~\ref{sec: ablation}): (\textbf{i}) \textbf{PPO}~\cite{schulman2017proximal} trains an imitation policy from scratch, following~\cite{peng2018deepmimic}. We experiment with both versions, with and without \textit{reference distillation}; (\textbf{ii}) \textbf{DAgger}~\cite{ross2011reduction} distills the student without RL fine-tuning, a process we refer to as \textit{policy distillation}.

\noindent\textbf{Implementation Details.} The control policies operate at 30 Hz and are trained using the \textit{Isaac Gym} simulator~\cite{makoviychuk2021isaac}. Teacher policies are implemented as MLPs with hidden layers of sizes 1024, 1024, and 512. The student policy is implemented as a three-layer Transformer encoder with 4 heads, a hidden size of 256, and a feed-forward layer of 512. The critics are also modeled as MLPs with the same architecture as the teacher policies.
To integrate the student policy with kinematic generators, including text-to-HOI~\cite{peng2023hoi} and future interaction prediction~\cite{xu2023interdiff}, we train these models using reference data distilled by the teacher policies from the OMOMO~\cite{li2023object} dataset, following the same train-test split as the student policy training. For the text-to-HOI model, we train it to generate 10 seconds of motion and use 24 generated samples for evaluation, while for future interaction prediction, the model generates 25 future frames given 10 past frames and we use 60 generated samples for evaluation. See Sec.~\ref{sec:training} of the supplementary.

\begin{figure}
    \centering
    \includegraphics[width=\columnwidth]{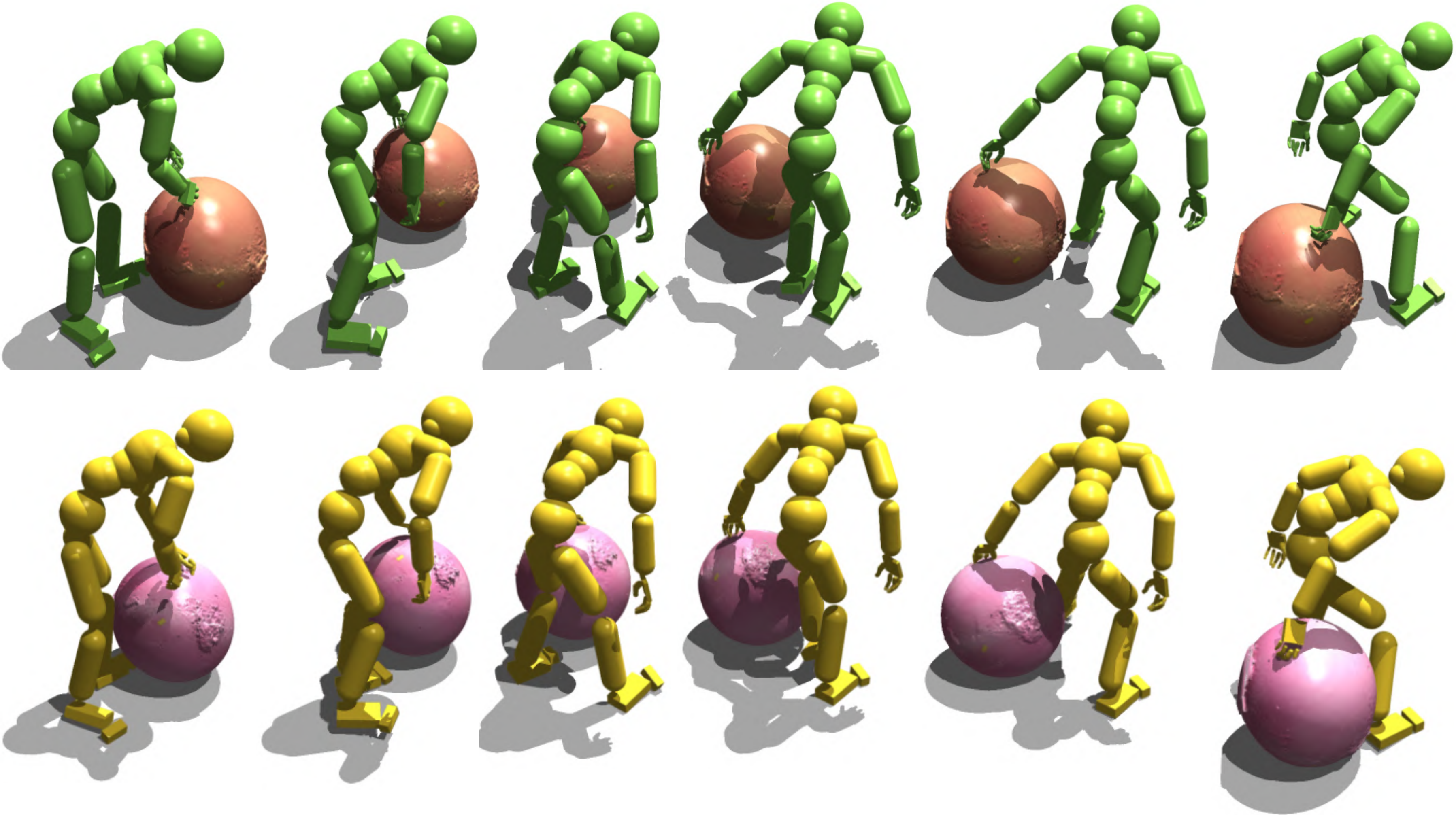}
    \caption{We recover plausible object rotations (bottom) that are challenging for motion capture due to the equivariant geometries of objects, which result in the object sliding on the ground (top).
    }
    \label{fig:obj_rot}
\end{figure}
\begin{table}
\centering
\resizebox{0.8\columnwidth}{!}{
\begin{tabular}{cccccc}
\toprule
Method  & Time$^\uparrow$  & $E_h$$^\downarrow$& $E_o$$^\downarrow$\\
\midrule
SkillMimic~\cite{wang2024skillmimic} &   12.2  &  7.2     &   13.4   \\
InterMimic (\textbf{Ours}) w/o IET   &  40.3 &     6.7    &    9.9   \\
InterMimic (\textbf{Ours}) w/o PSI   & 36.1   &     6.6   &    10.2 \\
InterMimic (\textbf{Ours})   &   \textbf{42.6}  &     \textbf{6.4}    &    \textbf{9.2}  \\
\bottomrule
\end{tabular}}
\caption{Quantitative comparison between the teacher policy from InterMimic and SkillMimic~\cite{wang2024skillmimic} to imitate data extracted from the BEHAVE~\cite{bhatnagar22behave} dataset involving a single subject interacting with yogamat. We ablate our proposed approach by removing interaction early termination and physical state initialization.}
\label{table:comparison}
\vspace{-0.5em}
\end{table}

\subsection{Quantitative Evaluation}\label{sec:quan}
\begin{table*}
\centering
\resizebox{\textwidth}{!}{
\begin{tabular}{cccccccccccccccccccccccc}
\toprule
\multirow{2}{*}{PPO}  & \multirow{2}{*}{\makecell{Reference \\ Distillation}}& \multirow{2}{*}{\makecell{Policy \\ Distillation}}& \multirow{2}{*}{Architecture} & \multicolumn{4}{c}{OMOMO-Train~\cite{li2023object}} & \multicolumn{4}{c}{OMOMO~\cite{li2023object}-Test} & \multicolumn{4}{c}{OMOMO~\cite{li2023object}-Test (w $\times 10$)} & \multicolumn{4}{c}{HOI-Diff~\cite{peng2023hoi}} & \multicolumn{4}{c}{InterDiff~\cite{xu2023interdiff}} \\
\cmidrule(lr){5-8}\cmidrule(lr){9-12}\cmidrule(lr){13-16}\cmidrule(lr){17-20}\cmidrule(lr){21-24}
& & && Succ.$^\uparrow$ & Time$^\uparrow$ & $E_h$$^\downarrow$ & $E_o$$^\downarrow$ & Succ.$^\uparrow$ & Time$^\uparrow$ & $E_h$$^\downarrow$ & $E_o$$^\downarrow$ & Succ.$^\uparrow$ & Time$^\uparrow$  & $E_h$$^\downarrow$ & $E_o$$^\downarrow$ & Succ.$^\uparrow$ & Time$^\uparrow$ & $E_h$$^\downarrow$ & $E_o$$^\downarrow$ & Succ.$^\uparrow$ & Time$^\uparrow$ & $E_h$$^\downarrow$ & $E_o$$^\downarrow$\\
\midrule
$\checkmark$ & $\times$ & $\times$ & \multirow{4}{*}{MLP} & 23.9 & 101.6 & 7.2 & 15.6 & 9.6 & 85.3 & 7.5 & 16.2 & 3.9 & 71.2  & \underline{7.5} & 17.9 & 0.0 & 0.0 & - & - & 6.7 & 11.7 & \textbf{6.2} & 16.4 \\
$\times$ & $\checkmark$ & $\checkmark$ & & 54.5 & 139.9 & 7.1 & 11.0 & 54.3 & 140.2 & 7.1 & \textbf{11.2} & 15.5 & 91.7  & 9.3 & \underline{13.7} & 4.2 & 84.8 & 10.1 & \textbf{9.7} &  65.0 & 27.4 & 7.5 & \underline{13.4} \\
$\checkmark$ & $\checkmark$ & $\times$ & & 71.7 & 152.8 & 8.9 & 12.7 & 91.6 & 173.7  & 8.5 & 13.2 & 45.8 & 127.6  & 9.1 & 14.9 & \underline{8.3} & \textbf{130.9} & 10.1 & 13.8 & 73.3 & 28.9 & 6.9 & 14.4 \\
$\checkmark$   & $\checkmark$ & $\checkmark$ & &\textbf{90.7} & \textbf{168.0} & \textbf{5.5} & \textbf{9.7} & \underline{95.5} & \underline{173.9} & \textbf{5.4} & {11.9} & \textbf{62.6} & \textbf{140.9} & \textbf{6.6} & {14.5} & \textbf{12.5} & \underline{121.4} & \underline{8.6} & \underline{12.1} & \underline{75.0} & \underline{29.1} & \textbf{6.2} & {13.5}\\
\midrule
$\checkmark$ & $\checkmark$ & $\checkmark$ & Transformer & \underline{88.8} & \underline{167.0} & \underline{6.0} & \underline{10.2} & \textbf{98.1} & \textbf{176.5} & \underline{5.9} & \underline{11.3} & \underline{56.8} & \underline{134.7} & \textbf{6.6} & \textbf{13.2} & \textbf{12.5} & {119.0} & \textbf{8.5} & {12.6} & \textbf{76.7} & \textbf{29.3} & \underline{6.4} & \textbf{13.3}\\
\bottomrule
\end{tabular}}
\caption{Quantitative evaluation of large-scale interaction imitation using OMOMO~\cite{li2023object}, kinematic generations from HOI-Diff~\cite{peng2023hoi}, and InterDiff~\cite{xu2023interdiff}. Additionally, we evaluate on test set when objects with weights ten times greater than those used during training.}
\label{table:student}
\end{table*}

Table~\ref{table:comparison} shows that the baseline struggles with MoCap imperfections, \eg, incorrect hand positioning, and thus results in clearly shorter tracking durations. In contrast, our method maintains reference tracking for longer durations and produces interactions that closely match the reference. Table~\ref{table:student} shows that our method consistently outperforms baselines in both training data imitation and out-of-distribution generalization, including interactions from the test set and from kinematic generations. We discuss the effectiveness of specific design choices in Sec.~\ref{sec: ablation}.

\subsection{Qualitative Evaluation}\label{sec:qual}
\begin{figure}
    \centering
    \includegraphics[width=\columnwidth]{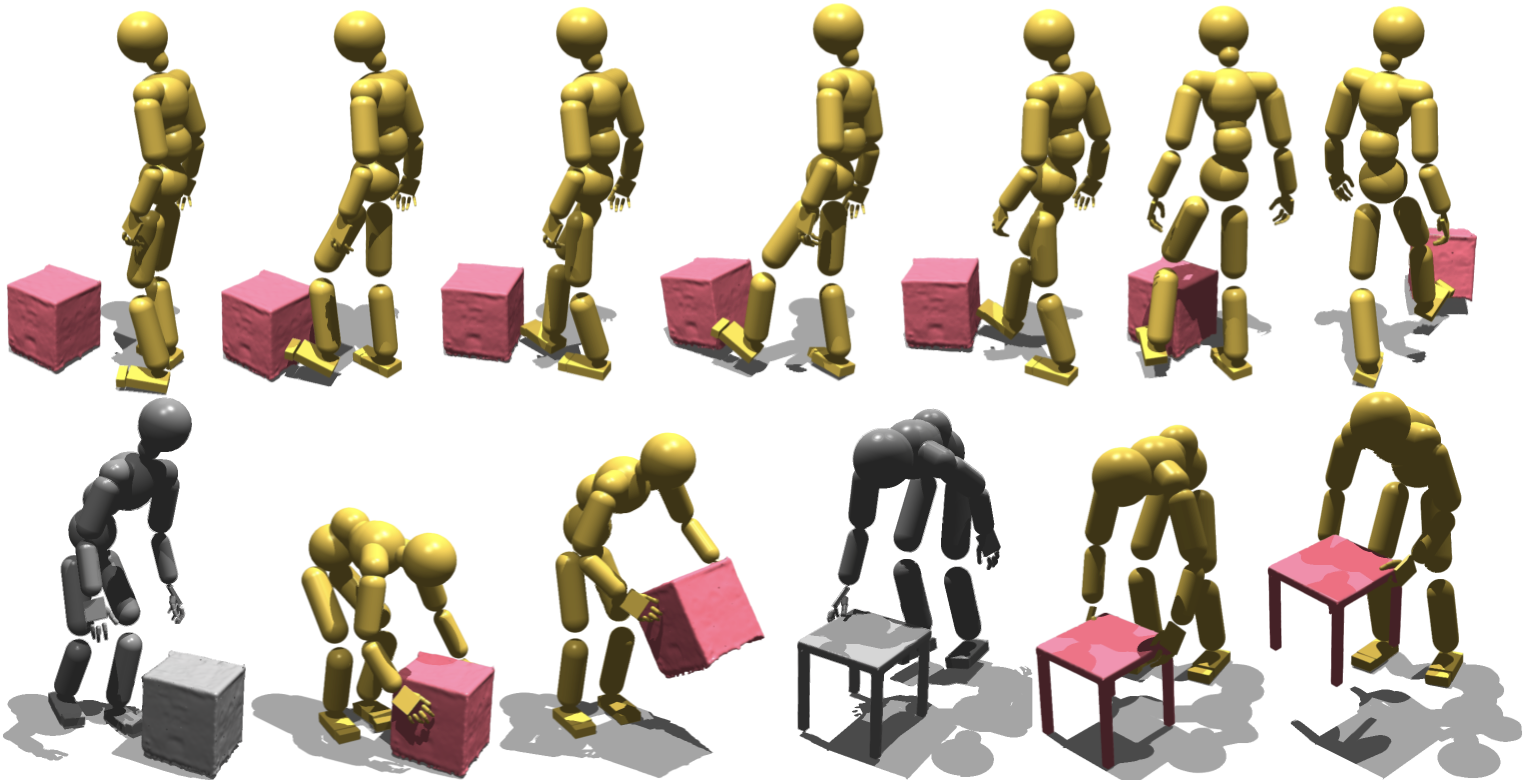}
    \caption{Zero-shot integration with a text-to-HOI model HOI-Diff~\cite{peng2023hoi} (\textbf{Top}), using `Kick the large box' as the prompt, and an interaction prediction model InterDiff~\cite{xu2023interdiff} (\textbf{Bottom}), where gray meshes are past states and colored illustrate future generations.
    }
    \label{fig:zero}
\end{figure}

\begin{figure}
    \centering
    \includegraphics[width=\columnwidth]{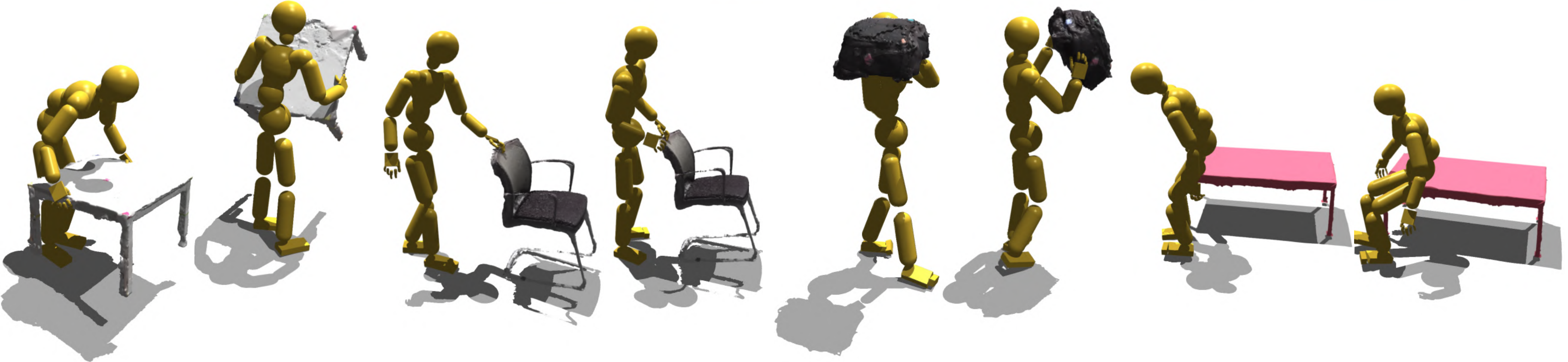}
    \caption{Zero-shot generalization of our student policy on novel objects from BEHAVE~\cite{bhatnagar22behave} and HODome~\cite{zhang2023neuraldome}.
    }
    \label{fig:zero2}
\end{figure}

Figure~\ref{fig:compare} shows a representative sequence from the experiment in Table~\ref{table:comparison}, illustrating how our teacher policy corrects interactions that PhysHOI~\cite{wang2023physhoi} fails to track robustly -- our method effectively withstands and corrects incorrect hand positioning and floating contacts in the reference. Beyond obvious errors, our method also rectifies the rotation of symmetric objects that MoCap inaccurately depicts as sliding along the ground, shown in Figure~\ref{fig:obj_rot}. Figure~\ref{fig:zero} presents additional examples that complement Figure~\ref{fig:teaser}, demonstrating how our approach integrates with kinematic generators for future interaction prediction and text-to-interaction synthesis. This zero-shot generalization extends to novel objects unseen during training (Figure~\ref{fig:zero2}), highlighting the effectiveness of our object geometry and contact-encoded representation, as well as the large-scale training.

\subsection{Ablation Study} \label{sec: ablation}

\noindent\textbf{Effectiveness of PSI and IET.}
We conduct an ablation study, as demonstrated in Table~\ref{table:comparison}, comparing the full approach to ``Ours w/o PSI''. The results validate that Physical State Initialization (PSI) is effective by mitigating inaccuracies in the motion capture data. We also observe reduced effectiveness without our interaction early termination, as training often spends rollouts on irrelevant periods.

\noindent\textbf{Effectiveness of Reference Distillation.} 
Compared to directly scaling imitation from MoCap with potential imperfections (line 1 in Table~\ref{table:student}), using references refined by the teacher policy (line 3) achieves consistently better performance on all metrics. The improvement is even more pronounced on the test set, where, without reference distillation, the policy struggles with unseen shapes, while retargeting by reference distillation eliminates the difficulty.

\noindent\textbf{Effectiveness of Joint PPO and DAgger Updates.} As shown in Table~\ref{table:student}, training a policy from scratch (line 3) or relying solely on policy distillation (DAgger, line 2) fails to achieve optimal performance. While supervised skill learning lays the groundwork, additional PPO fine-tuning is crucial for resolving conflicts among teacher policies. This is important because our subject-based clustering may not effectively distinguish between different interaction patterns, and ambiguity arises when multiple teachers produce different actions for similar motions.

\noindent\textbf{Effectiveness of Transformer for Policy Learning.}  
From Table~\ref{table:student}, we see that using a Transformer policy (line 5) outperforms MLP-based approaches, particularly on the test set and out-of-distribution cases generated by the kinematic model. We attribute this to the Transformer's inductive bias in sequential modeling and its capacity to incorporate longer-term observations, enabling it to handle complex spatio-temporal dependencies more effectively.

\section{Conclusion}
In this work, we introduce a framework for synthesizing realistic human-object interactions that are both physically grounded and generalizable. Unlike previous methods, our approach leverages a rich repository of imperfect MoCap data to facilitate the learning of various interaction skills across a wide variety of objects. To address inaccuracies in the MoCap data, we propose contact-guided rewards and optimize trajectory collection, enabling teacher policies to recover missing physical details in the original data.
These teacher policies are used to train student policies within a distillation framework that combines policy distillation and reference distillation, thus enabling efficient skill scaling. Our approach shows zero-shot generalizability, which effectively bridges the gap between imitation and generative capabilities by integrating with kinematic generation. We believe that this framework can be adapted for whole-body loco-manipulation for real-world robots, enabling them to handle objects with human-like dexterity and nuance.
\newpage
\paragraph{Acknowledgments.} We thank Wei Yang, Yu-Wei Chao, Arsalan Mousavian, Ankur Handa, Samuel Schulter, Morteza Ziyadi, and Xiaohan Fei for valuable discussions. This work was supported in part by NSF Grant 2106825, NIFA Award 2020-67021-32799, the Amazon-Illinois Center on AI for Interactive Conversational Experiences, the Toyota Research Institute, the IBM-Illinois Discovery Accelerator Institute, and Snap Inc. This work used computational resources, including the NCSA Delta and DeltaAI and the PTI Jetstream2 supercomputers through allocations CIS230012, CIS230013, and CIS240311 from the Advanced Cyberinfrastructure Coordination Ecosystem: Services \& Support (ACCESS) program, as well as the TACC Frontera supercomputer and Amazon Web Services (AWS) through the National Artificial Intelligence Research Resource (NAIRR) Pilot.

{
    \small
    \bibliographystyle{ieeenat_fullname}
    \bibliography{main}
}

\clearpage
\setcounter{page}{1}
\maketitlesupplementary

\setcounter{table}{0}
\renewcommand{\thetable}{\Alph{table}}
\renewcommand*{\theHtable}{\thetable}
\setcounter{figure}{0}
\renewcommand{\thefigure}{\Alph{figure}}
\renewcommand*{\theHfigure}{\thefigure}
\setcounter{section}{0}
\renewcommand{\thesection}{\Alph{section}}
\renewcommand*{\theHsection}{\thesection}

\begin{strip}\centering
\vspace{-2em}
\includegraphics[width=\textwidth]{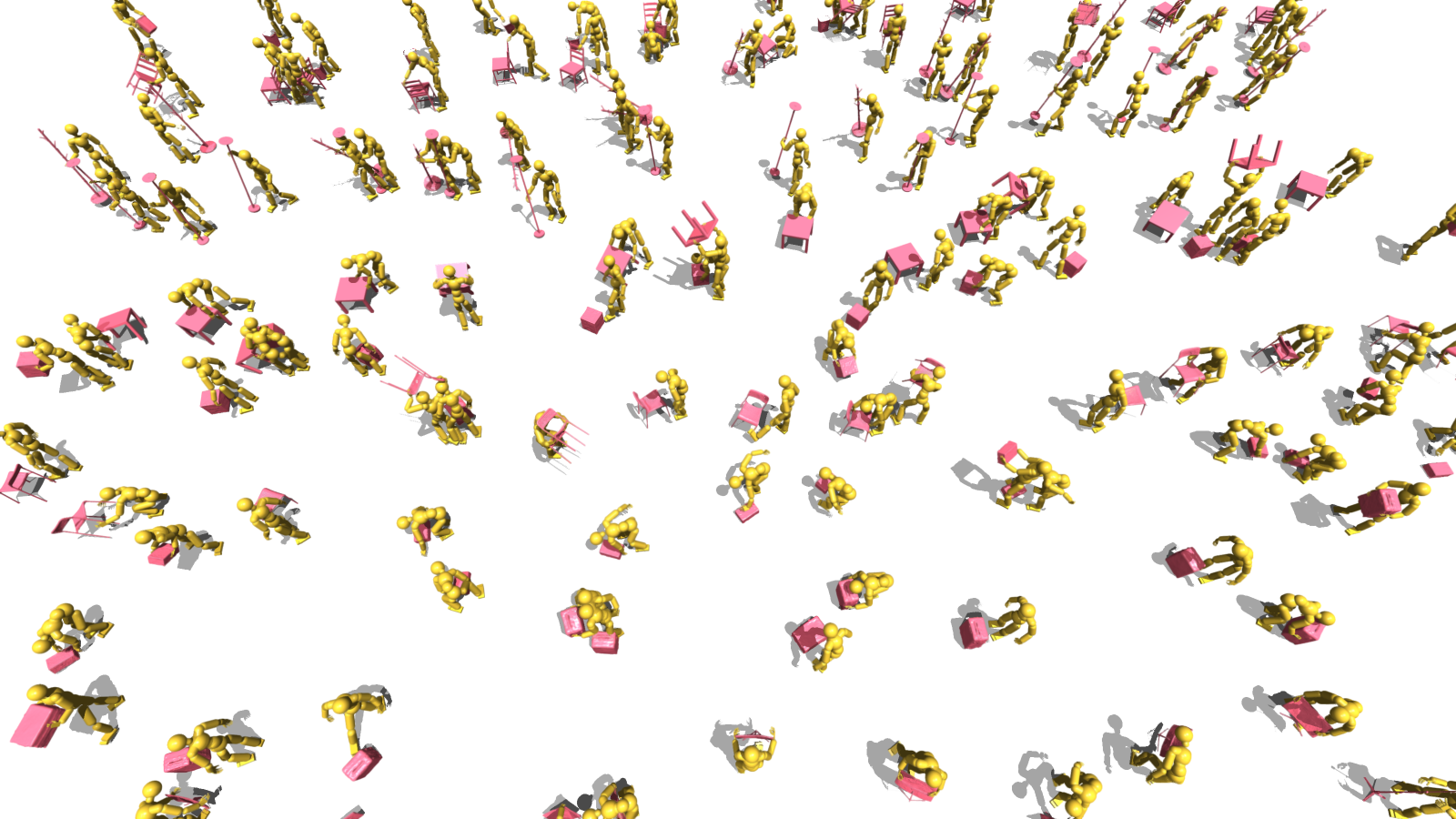}
\captionof{figure}{InterMimic enables simulated humans to perform physical interactions, featuring scalable skill learning covering diverse objects.
\label{fig:teaser_supp}}
\vspace{-1em}
\end{strip}

\noindent In this supplementary, we provide additional method details and experimental setups:
\begin{enumerate}[label=(\textbf{\roman*})]
    \item \textbf{Demo Video.} A demonstration video (with a screenshot in Figure~\ref{fig:teaser_supp}) is provided at \href{https://sirui-xu.github.io/InterMimic/assets/demo.mp4}{demo.mp4}, as described in Sec.~\ref{sec:demo}.
    \item \textbf{Simulation Setup.} The environment configuration for physical HOI simulations is introduced in Sec.~\ref{sec:phys_para}.
    \item \textbf{Reference Contact Labels.} Additional information on obtaining the reference contact label \(\hat{\boldsymbol{c}}_t\) is detailed in Sec.~\ref{sec:repre_supp}.
    \item \textbf{Reward Formulation.} A comprehensive explanation of the reward design is provided in Sec.~\ref{sec:reward}.
    \item \textbf{Physical State Initialization \& Interaction Early Termination.} Further insights into these mechanisms are discussed in Sec.~\ref{sec:psi_supp}.
    \item \textbf{Implementation Details.} This section (Sec.~\ref{sec:training}) covers reframing our method for interaction prediction and text-guided interaction generation, as well as translating MoCap interactions into humanoid robot skills.
    \item \textbf{Additional Experiments.} Sec.~\ref{sec:add_exp} presents further qualitative results and analyzes failure cases.
    \item \textbf{Limitations and Societal Impact.} Finally, we examine the limitations of InterMimic and its potential societal implications in Sec.~\ref{sec:discuss}.
\end{enumerate}

\noindent We will release the code for this project at \href{https://sirui-xu.github.io/InterMimic}{our webpage}.

\etocdepthtag.toc{mtappendix}
\etocsettagdepth{mtchapter}{none}
\etocsettagdepth{mtappendix}{subsection}
{
  \hypersetup{
    linkcolor = black
  }
  \tableofcontents
}

\section{Demo Video} \label{sec:demo}

In addition to the qualitative results presented in the main paper, we provide a demo video \href{https://sirui-xu.github.io/InterMimic/assets/demo.mp4}{(demo.mp4)} for more detailed visualizations of the tasks, further illustrating the efficacy of our approach. The demo video conveys the following key points:

\begin{enumerate}[label=(\textbf{\roman*})]
\item Our teacher policy can imitate highly dynamic and long-term interactions, both of which are inherently challenging.

\item We visualize the effectiveness of our teacher policy in \textit{HOI retargeting}. Given MoCap references for humans, we successfully transfer these tasks to a humanoid robot, tolerating embodiment differences.

\item Our method corrects errors in reference interactions, addressing contact penetration, floating, and jittering issues. This demonstrates how teacher-based reference distillation can provide cleaner data for student policy training.

\item The baseline method PhysHOI~\cite{wang2023physhoi} \textit{fails} on sequences our approach successfully imitates, complementing Figure~\ref{fig:compare} in the main paper.

\item Our student policy exhibits strong scalability, effectively learning from hours of data across diverse objects and interaction skills.

\item The framework grants the student policy \textit{zero-shot} generalizability, enabling direct application to text-to-HOI, interaction prediction, and interactions with new skills or objects -- even multiple objects not present in the training set.
\end{enumerate}

\begin{figure}
    \centering
    \includegraphics[width=\columnwidth]{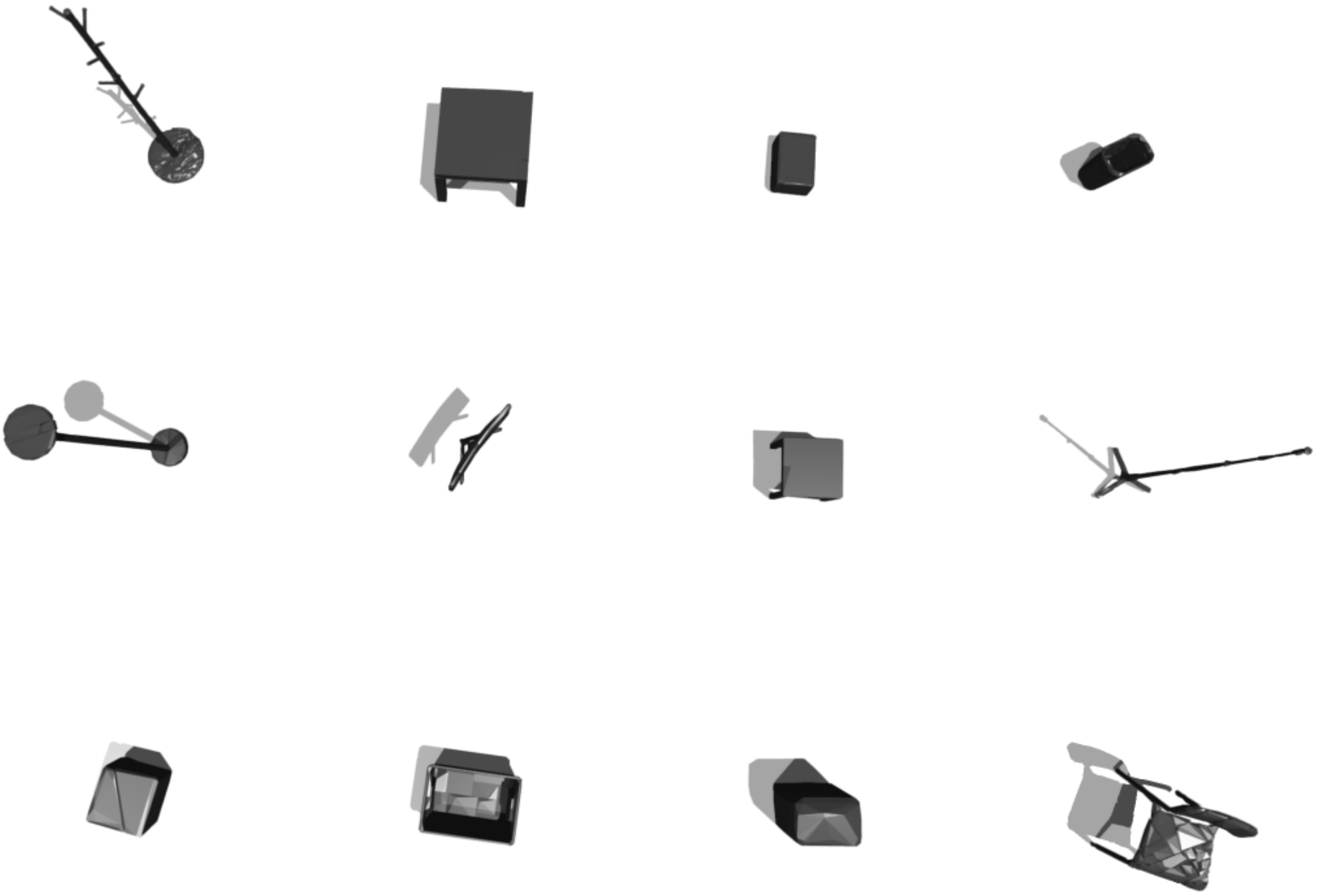}
    \caption{Visualization of the objects from OMOMO~\cite{li2023object}, each decomposed into 64 convex hulls for simulation.
    }
    \label{fig:object}
\end{figure}

\section{Setup of Physical Interaction Simulation} \label{sec:phys_para}
The reference data represent humans using the SMPL models~\cite{MANO, SMPL-X:2019}. For simulation, we convert these models into box and cylindrical rigid bodies following~\cite{ luo2023perpetual}. Objects are also converted into simulation models through convex decomposition, as illustrated in Figure~\ref{fig:object}. We summarize the physics parameters for our task in Table~\ref{tab:physics_hyper}. We follow the physics parameters for the human as specified in~\cite{wang2023physhoi, wang2024skillmimic}, with the exception of the specialized range of motion (RoM) for hands, detailed in Table~\ref{tab:rom}. Our range of motion (RoM) setting is biologically inspired: finger flexion and extension (bending and straightening) are fully activated. However, unlike the real human, the abduction and adduction of the Metacarpophalangeal (MCP) joint are constrained to minimize the risk of finger interpenetration, in the absence of the correct reference hand pose for guidance. The rationale for these RoM settings is discussed in Sec.\ref{sec:teacher} of the main paper and Sec.\ref{sec:hand} of the supplementary.
\begin{table}
  \begin{tabular}{l|l}
    \toprule
    Hyperparameter & Value \\
    \midrule
    Sim $dt$ & $1/60$s\\
    Control $dt$ & $1/30$s\\
    Number of envs & 8192 \\
    \midrule
    Number of substeps & 2 \\
    Number of pos iterations & 4 \\
    Number of vel iterations & 0 \\
    Contact offset & 0.02 \\
    Rest offset & 0.0 \\
    Max depenetration velocity & 100 \\
    \midrule
    Object \& ground restitution & 0.7 \\
    Object \& ground friction & 0.9 \\ 
    Object density & 200 \\
    Object max convex hulls & 64 \\
  \bottomrule
\end{tabular}\caption{Simulation hyperparameters used in Isaac Gym~\cite{makoviychuk2021isaac}.}
\label{tab:physics_hyper}
\end{table}

\begin{table}
  \begin{tabular}{l|l|l}
    \toprule
    Joint & x-dim & y \& z-dim \\
    \midrule
    MCP & $[-55.625^\circ, 55.625^\circ]$ & \multirow{3}{*}{$[-5.625^\circ, 5.625^\circ]$} \\
    PIP & $[-55.625^\circ, 55.625^\circ]$ &  \\
    DIP & $[-5.625^\circ, 90.000^\circ]$ &  \\
    \midrule
    CMC &$[-55.625^\circ, 55.625^\circ]$&$[-55.625^\circ, 55.625^\circ]$\\
    MCP &$[-5.625^\circ, 5.625^\circ]$&$[-5.625^\circ, 5.625^\circ]$\\
    IP &$[-5.625^\circ, 90.000^\circ]$&$[-5.625^\circ, 5.625^\circ]$\\
  \bottomrule
\end{tabular}
\caption{We constrain the Range of Motion (RoM) for the joints in the index, middle, ring, and pinky fingers including: the MCP (Metacarpophalangeal) joint where the finger meets the hand, the PIP (Proximal Interphalangeal) joint as the middle joint, and the DIP (Distal Interphalangeal) joint closest to the fingertip. For the thumb, we consider the CMC (Carpometacarpal) joint at the base in the palm, the MCP connecting the thumb to the hand, and the IP (Interphalangeal) joint within the thumb. The coordinates for describing these RoMs are based on the human model from~\cite{luo2023perpetual}.}
\label{tab:rom}
\end{table}
\section{Reference Contact} \label{sec:repre_supp}
In this section, we detail how we extract the reference contact that formulates the state and the reward as discussed in Sec.~\ref{sec:state} of the main paper. 
One solution involves loading the HOI data into the simulation, replaying the data, and using the force detector in Isaac Gym~\cite{makoviychuk2021isaac} to identify contact, as suggested by~\cite{wang2023physhoi}. However, this approach is ineffective for imperfect MoCap data; for instance, the force detector fails to capture contact when floating artifacts occur. To address this limitation, we propose solutions tailored differently for teacher and student training:

\noindent\textbf{Reference contact for the student.} We query the force detector from distilled reference in the simulation rather than from MoCap data replay, as the teacher policy is capable of correcting artifacts.

\noindent\textbf{Reference contact for teachers.} To account for contact distance variances, we determine reference contact based on inferred dynamics from kinematics, as outlined below.

\subsection{Inferring Reference Dynamics} \label{sec:contact_lab}
By analyzing the object’s acceleration over time, we can approximate external forces without depending on simulated dynamics. We assume human-object interaction occurs if any of these conditions hold:
(\textbf{i}) The object is airborne, but its acceleration deviates significantly from gravitational acceleration, indicating that an external force, \eg, human interaction is acting upon it.
(\textbf{ii}) The object is on the ground but not static, and its acceleration significantly differs from what is expected due to friction alone, suggesting additional force input.
(\textbf{iii}) The minimum distance between the human and object vertices is below 0.01 meters.

When any condition is met, we define the contact threshold $\sigma$ as the minimum distance between the human and object vertices, plus 0.005 meters. This adaptive threshold is essential for accommodating contact distance variations in the ground truth MoCap data. For example, the contact promotion marker is defined as $\hat{\boldsymbol{c}}_b[i] = \|\hat{\boldsymbol{d}}[i]\| < \sigma$, where $i$ is the index of human rigid bodies. We integrate $\hat{\boldsymbol{c}}_b$ into the contact promotion reward $R^c_b$, as introduced in Sec.~\ref{sec:teacher} of the main paper and detailed in Sec.~\ref{sec:contact_reward} of supplementary. $\hat{\boldsymbol{d}}$ is the joint-to-object vectors as defined in Sec.~\ref{sec:state}.

\section{Additional Details on Reward} \label{sec:reward}
In this section, we provide further details about the reward function used for policy training. Specifically, we describe how we balance the components of the embodiment-aware reward, formulate the contact and energy rewards, address hand interaction recovery, and explain the process of integrating all rewards into a unified scalar.

\subsection{Embodiment-Aware Reward}
We formulate the weight \(\boldsymbol{w}_d\), introduced in Sec.~\ref{sec:teacher} of the main paper, for balancing the embodiment-aware reward:
\begin{equation}
    \boldsymbol w_d[i] = 0.5 \times \frac{1 / \|\boldsymbol{d}[i]\|^2}{\sum_{i} 1 / \|\boldsymbol{d}[i]\|^2} + 0.5 \times \frac{1 / \|\hat{\boldsymbol{d}}[i]\|^2}{\sum_{i} 1 / \|\hat{\boldsymbol{d}}[i]\|^2},
    \label{eq:edge_weighting_function}
\end{equation}
where \(i\) is the joint index, and \(\boldsymbol{d}\) and \(\hat{\boldsymbol{d}}\) are vectors from the human joint to the object surface for simulation and reference, respectively, as defined in Sec.~\ref{sec:state} of the main paper. The value $\|\boldsymbol{d}[i]\|^2$ and $\|\hat{\boldsymbol{d}}[i]\|^2$ are clipped by a small positive value to prevent division by zero.

Our joint position and rotation tracking rewards, \(R^h_p\) and \(R^h_{\theta}\), include both body and hand joints, even for imitating datasets such as~\cite{bhatnagar22behave,li2023object} which present hands always in flat or mean poses. This encourages hands to maintain a reasonable default pose when the contact reward is not activated.

\subsection{Contact Reward} \label{sec:contact_reward}
The contact promotion cost function $E^{c}_b$ is designed to encourage highly probable contact, as highlighted by the red regions in Figure~\ref{fig:contact_label}(i) of the main paper. This reward utilizes the adaptive contact marker \(\hat{\boldsymbol{c}}_b\), described in Sec.~\ref{sec:contact_lab},
\begin{align}
E^{c}_b = \sum\|\hat{\boldsymbol{c}}_b - \boldsymbol{c}\| \odot \hat{\boldsymbol{c}}_b,
\end{align}
where $\boldsymbol{c}$ is the simulated contact extracted from the force detected, as introduced in Sec.~\ref{sec:state} of the main paper.

Contact penalties, applied to the blue regions in Figure~\ref{fig:contact_label}(i) of the main paper, are defined using a larger and fixed threshold of \(\sigma_p = 0.1\). Specifically, \(\hat{\boldsymbol{c}}_p[i] = (\|\hat{\boldsymbol{d}}[i]\| > \sigma_p) \land \neg \hat{\boldsymbol{c}}_g[i]\), where $\|\hat{\boldsymbol{d}}[i]\|$ is the distance between joint $i$ and the object surface in the reference interaction as defined in Sec.~\ref{sec:state} of the main paper, and the negation $\neg$ of \(\hat{\boldsymbol{c}}_g[i]\) indicates the rigid body part $i$ that is not in contact with the ground. The cost of penalty is then calculated as:
\begin{align}
E^c_p = \sum\|\boldsymbol{c}\| \odot \hat{\boldsymbol{c}}_p.
\end{align}

\subsection{Hand Interaction Recovery}\label{sec:hand}
Our hand contact guidance is defined as:
\begin{align}
E^c_h &= \sum\|\boldsymbol{c}^{\mathrm{lhand}} - \hat{\boldsymbol{c}}^{\mathrm{lhand}}\| \odot \hat{\boldsymbol{c}}^{\mathrm{lhand}} \\ &+ \|\boldsymbol{c}^{\mathrm{rhand}} - \hat{\boldsymbol{c}}^{\mathrm{rhand}}\| \odot \hat{\boldsymbol{c}}^{\mathrm{rhand}},
\end{align}
where \(\boldsymbol{c}^{\mathrm{lhand}}\) and \(\boldsymbol{c}^{\mathrm{rhand}}\) represent contact labels for rigid body components of the hands. The reference contact markers, \(\hat{\boldsymbol{c}}^{\mathrm{lhand}}\) and \(\hat{\boldsymbol{c}}^{\mathrm{rhand}}\), are defined when any hand vertices are within an adaptive threshold distance $\sigma$ to the objects, as described in Sec.~\ref{sec:contact_lab} of supplementary. 
To avoid overly aggressive hand contact that could lead to unrealistic poses, we impose range of motion constraints for the hand, as shown in Table~\ref{tab:rom}, ensuring that RL-explored grasping remains biologically realistic.

\subsection{Energy Reward} \label{sec:energy}
We define the energy cost as $E^{e}_h = \sum\|\boldsymbol{a}_h\|$, $E^e_{o} = \sum\|\boldsymbol{a}_o\|$, and $E^e_{c} = \max\|\boldsymbol{f}\|$,
where \( \boldsymbol{a}_h \) represents the acceleration of human joints, \( \boldsymbol{a}_o \) the object's acceleration, and \( \boldsymbol{f} \) the force detected on human rigid bodies. Applying them penalizes abrupt contact and promotes smooth interactions.

\subsection{Reward Aggregation}
We define the weights for each cost function, including \(E^h_p\), \(E^h_{\theta}\), \(E_d\), \(E^o_p\), and \(E^o_{\theta}\), as described in Sec.~\ref{sec:teacher} of the main paper, along with \(E^c_b\), \(E^c_p\), \(E^c_h\), \(E^e_h\), \(E^e_o\), and \(E^e_c\) detailed in supplementary as \((\lambda^h_p, \lambda^h_{\theta}, \lambda_d, \lambda^o_p, \lambda^o_\theta, \lambda_{c_b}, \lambda_{c_p}, \lambda_{c_h}, \lambda^h_e, \lambda^o_e, \lambda^f_e)\). The final aggregated reward is computed as:
\(
R = \exp( -\lambda^h_{\theta} E^h_{\theta} - \lambda^h_p E^h_p - \lambda^o_{\theta} E^o_{\theta} - \lambda^o_p E^o_p - \lambda_d E_d - \lambda_{c_b} E^c_b - \lambda_{c_p} E^c_p - \lambda_{c_h} E^c_h - \lambda^h_e E^e_h - \lambda^o_e E^e_o - \lambda^f_e E^e_c ).
\), following a multiplication of the exponential structure, as suggested in~\cite{won2020scalable,park2019learning}.

\section{Additional Details on Trajectory Collection} 
\label{sec:psi_supp}

\subsection{Interaction Early Termination}
In Sec.~\ref{sec:teacher} of the main paper, we introduce the termination conditions defined for human-object interaction. Additionally, we use three conditions general for single human imitation as follows:
(\textbf{i}) The joints are, on average, more than 0.5 meters from their reference.
(\textbf{ii}) The root joint is under the height of 0.15.
(\textbf{iii}) The episode ends after 300 frames, as the maximum episode length (also specified in Table~\ref{tab:ppo_hype}).

\subsection{Physical State Initialization}
\begin{figure}
    \centering
    \includegraphics[width=\columnwidth]{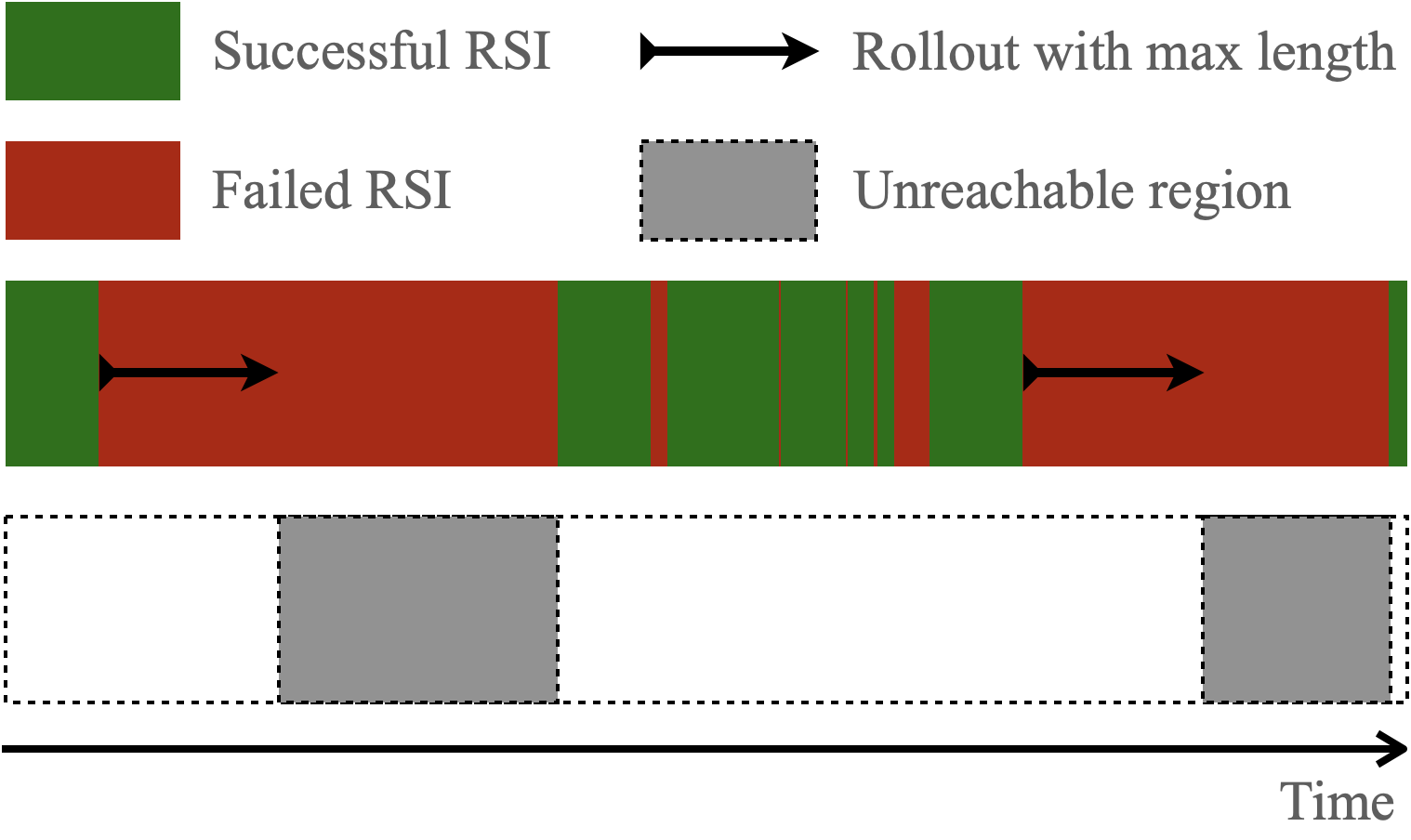}
    \caption{A sanity check on why Reference State Initialization (RSI)~\cite{peng2018deepmimic} can fail: we use a bar representing the reference interaction sequence that the policy imitates, where red regions indicate that initializing in those regions leads to immediate failure, while green regions signify that successful initialization is possible. There may be periods, shown as two gray blocks, where the policy cannot collect trajectories for updates (\ie, unreachable regions), as the successful rollout cannot cover large failed RSI region given the fixed length of the rollout. In real scenarios, rollouts can be suboptimal and terminated prematurely, preventing the policy from collecting sufficient trajectories for challenging periods that extend beyond the boundaries illustrated by the gray blocks.
    }
    \label{fig:psi_sanity}
\end{figure}

\noindent\textbf{Limitations of RSI.} Figure~\ref{fig:psi_sanity} illustrates why Reference State Initialization (RSI)~\cite{peng2018deepmimic} is suboptimal for interaction imitation with imperfect MoCap data. In single-person MoCap scenarios, where failures are less frequent, RSI performs well; however, in the presence of MoCap errors, RSI leads to reduced experience collection, ultimately undermining performance.

\noindent\textbf{Does Interaction Early Termination Help?} While early termination can filter out poor initial states, excessive initialization failures lead to frequent simulation resets that significantly slow down training. Consequently, the agent spends more time restarting simulations rather than engaging in productive learning.

\noindent\textbf{Step-by-step details} to complement Sec.~\ref{sec:teacher} of the main paper:
(\textbf{i})
PSI begins by creating an initialization buffer that stores a collection of reference states from motion capture data and simulation states from previous rollouts. This buffer is used to select initialization states for future rollouts.
(\textbf{ii})
For each new rollout, an initialization state is randomly selected from the buffer.
(\textbf{iii})
Using the current policy, the agent performs rollouts in the simulation environment by taking actions, transitioning through states, and receiving rewards.
(\textbf{iv})
After each rollout, the collected trajectories are evaluated based on their expected discounted rewards to update the critic network. Trajectories with expected rewards above a defined threshold are added to the PSI buffer, while older or lower-quality trajectories are removed to maintain the buffer's size and quality.
We apply PSI in a sparse manner to enhance training efficiency, with a probability of 0.005 for updating the buffer for each rollout.

\section{Additional Implementation Details}\label{sec:training}
\begin{figure}
    \centering
    \includegraphics[width=\columnwidth]{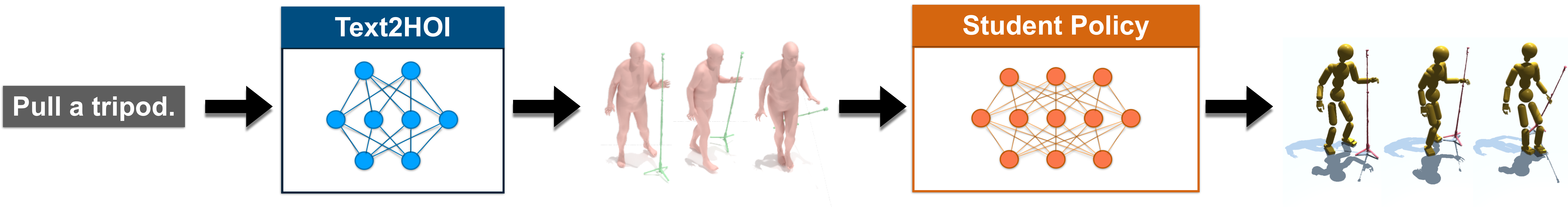}
    \caption{Overview of integrating HOI-Diff~\cite{peng2023hoi} with InterMimic to perform text-guided interaction generation, \ie, generating interaction sequences based on text input.
    }
    \label{fig:text2hoi}
\end{figure}

\begin{figure}
    \centering
    \includegraphics[width=\columnwidth]{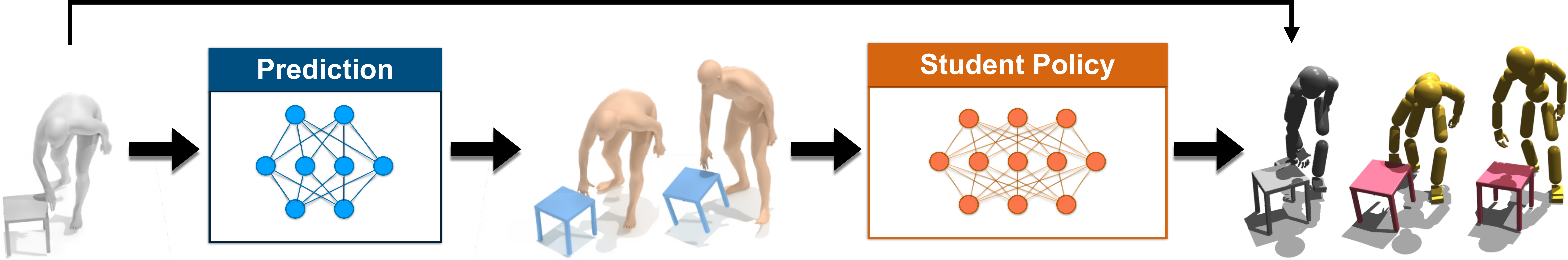}
    \caption{Overview of integrating InterDiff~\cite{xu2023interdiff} with InterMimic to perform interaction prediction, \ie, generating future interactions based on past interaction frames.
    }
    \label{fig:interdiff}
\end{figure}

In Figures~\ref{fig:text2hoi} and \ref{fig:interdiff}, we illustrate the framework that integrates the kinematic generators with our InterMimic -- let the policy use the kinematic output as the input reference to imitate. Table~\ref{tab:ppo_hype} lists the hyperparameters used during the PPO~\cite{schulman2017proximal}. The weights for the reward function \((\lambda^h_p, \lambda^h_{\theta}, \lambda_d, \lambda^o_p, \lambda^o_\theta, \lambda_{c_b}, \lambda_{c_p}, \lambda_{c_h}, \lambda^h_e, \lambda^o_e, \lambda^f_e)\) are set as $(30, 2.5, 5, 0.1, 5, 5, 5, 3, 2\times10^{-5}, 2\times10^{-5}, 10^{-9})$. 

For evaluation on the OMOMO~\cite{li2023object} dataset, we use Subject 9 as the base model, with teacher policies retargeting interactions from other subjects into this base.

Similar to existing motion imitation approaches~\cite{peng2018deepmimic}, we use API in Isaac Gym~\cite{makoviychuk2021isaac} to initialize the first frame to match the first reference frame -- whether it comes from MoCap or kinematic generation methods. The subsequent sequence is then simulated based on the starting frame.

For learning interaction skills on a humanoid robot~\cite{unitreeg1,inspire} from SMPL-X~\cite{SMPL-X:2019} data, we bypass external retargeting and directly learn, highlighting our framework’s integrated ability for both retargeting and imitation. Note that we model each Inspire hand with 12 actuators using PD control, without accounting for the mimic joint present in the actual setup, which could be inapplicable in real deployment. Due to the embodiment gap, the humanoid cannot be initialized to match the first SMPL-X frame. Thus, we adopt a two-stage approach: during the first 15 frames, the policy learns to stand and approach the reference’s initial pose, establishing a basis for subsequent tracking. Afterward, the policy transitions to track the reference. We rewrite the position and rotation rewards for the robot’s joints mapped to SMPL-X joints. We do not use the contact reward as we disable the self-collision, since the human reference now cannot ensure proper collision constraints for the humanoid robot. To mitigate the impact of contact artifacts in MoCap data without relying on a contact reward, we leverage teacher distillation references for training.

For interactions involving multiple objects, our framework remains unchanged except for the state and reward components related to the objects, such as \(\{{\boldsymbol \theta}_t^o,  \boldsymbol p_t^o, {\boldsymbol \omega}_t^o,  \boldsymbol v_t^o\}\), \(\boldsymbol{d}_t\), and the rewards \(R^o_p\), \(R^o_{\theta}\), and \(R_d\), which now include multiple components to represent multiple objects.

\begin{table}[h]
  \begin{tabular}{l|l}
    \toprule
    Hyperparameters & value \\
    \midrule
    Action distribution & 153D Continuous\\
    Discount factor $\gamma$    & 0.99\\
    Generalized advantage estimation $\lambda$    & 0.95\\
    Entropy regularization coefficient & 0.0\\
    Optimizer & Adam~\cite{Adam} \\
    Learning rate (Actor) & 2e-5 \\
    Learning rate (Critic) & 1e-4 \\
    Minibatch size & 16384 \\
    Horizon length $H$ & 32 \\
    Action bounds loss coefficient & 10 \\
    Maximum episode length & 300 \\
  \bottomrule
\end{tabular}
\caption{Hyperparamters for training teacher and student policies.}
\label{tab:ppo_hype}
\end{table}

\section{Additional Experiemental Results}\label{sec:add_exp}
In this section, we introduce experimental results that are not included in the main paper due to space limit. 

\begin{figure}
    \centering
    \includegraphics[width=\columnwidth]{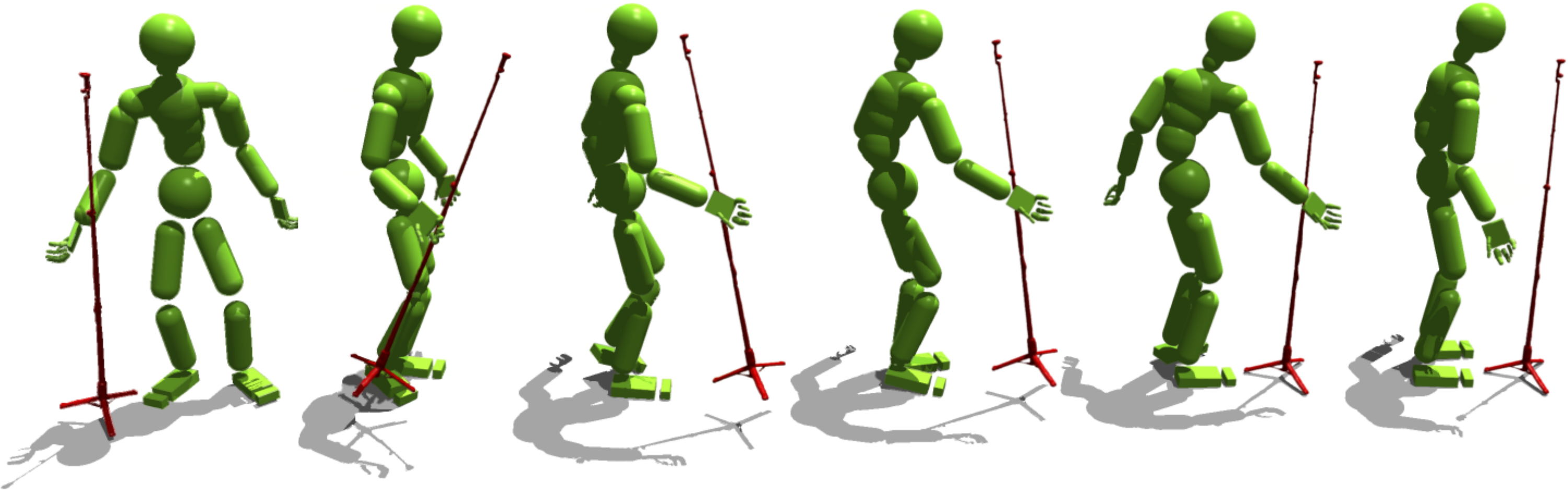}
    \caption{For certain reference from OMOMO~\cite{li2023object}, the hand is incorrectly flipped, which leads to the failure of the teacher policy. We exclude such data when training the student policy.
    }
    \label{fig:error}
\end{figure}

\noindent\textbf{Failure Cases.} In Figure~\ref{fig:error}, we illustrate an example where our teacher policies fail to perform successful imitation. Despite the strong adaptability of our policies, as demonstrated in Figures~\ref{fig:teaser} and \ref{fig:obj_rot}, where they effectively correct reference errors, there are limitations when encountering too many errors. Since the reward design inherently prioritizes reference tracking, excessive errors in the reference inevitably result in failures.

\begin{figure}
    \centering
    \includegraphics[width=\columnwidth]{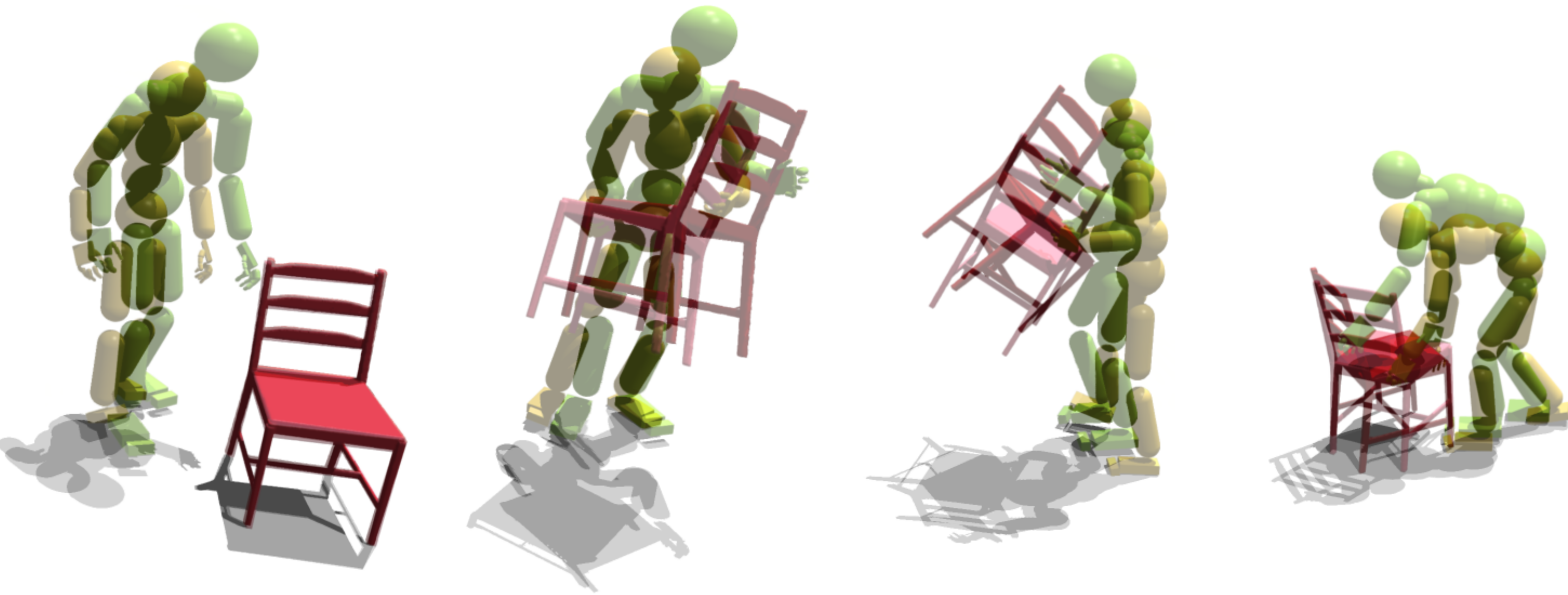}
    \caption{Comparison between the reference interaction (human shown in green) and the simulated interaction (human shown in yellow) demonstrates that, despite the different body shapes, the simulated human driven by InterMimic successfully accomplishes the same task with different trajectories, highlighting the effectiveness of our imitation as retargeting.
    }
    \label{fig:retargeting}
\end{figure}

\noindent\textbf{HOI Retargeting.} Figure~\ref{fig:retargeting} shows that teacher policies, trained on reference data for a specific body shape, can successfully drive a human model with a body shape different from the reference in the simulator to accomplish the same task, albeit with slightly varied trajectories. This result highlights the effectiveness of our design, which integrates retargeting into interaction imitation.

\section{Discussion} \label{sec:discuss}
\noindent\textbf{Limitations and Future Work.} One limitation, as discussed in Sec.~\ref{sec:add_exp} and illustrated in Figure~\ref{fig:error}, is that our method struggles to fully correct MoCap data with significant errors. However, it also underscores a strength of our teacher-student framework: teacher policies filter out data that are too corrupted to imitate, allowing the student policy to concentrate on learning from viable samples and avoid wasting training effort on unrecoverable data.

The policy sometimes results in unnatural object support, where the human produces penetration rather than relying on friction. While we mitigate this issue by setting a high maximum depenetration velocity in simulation (See Table~\ref{tab:physics_hyper}) and applying a contact-based energy (See Sec.~\ref{sec:energy}) to discourage large forces that could cause penetration, it does not entirely solve the problem. A potential solution could involve using a signed distance-based penetration score as a criterion for early termination.

The hand interaction recovery method is effective for the tasks explored in this paper. For tasks requiring dexterity with detailed finger motions, its benefits may be limited.

Additionally, while our method demonstrates good scalability by effectively training on hours of MoCap data involving different objects and generalizing to unseen skills and object geometries, its performance could be further improved with a larger dataset. Incorporating more diverse objects~\cite{xie2024intertrack} would likely further enhance InterMimic's zero-shot generalization capabilities.

\noindent\textbf{Potential Negative Societal Impact.}
Our approach has the potential to generate vivid human-object interaction sequences, which, if misused, could lead to negative societal impacts, with the risk of creating misleading content by depicting individuals in fabricated scenarios. However, our model is designed with privacy in mind -- it employs an abstract representation, using simple geometric shapes like boxes and cylinders to depict different parts. This abstraction reduces the inclusion of personally identifiable features, making it less likely for our synthesized data to be misused in ways that compromise individual identities.
\end{document}